\documentclass{article}
\usepackage{iclr2027_conference,times}

\usepackage[utf8]{inputenc}
\usepackage[T1]{fontenc}
\usepackage{amsmath,amssymb}
\usepackage{booktabs}
\usepackage{nicefrac}
\usepackage{microtype}
\usepackage{xcolor}
\usepackage{graphicx}
\usepackage{float}
\usepackage{url}
\usepackage{hyperref}
\usepackage{caption}
\usepackage{wrapfig}
\usepackage{fontawesome5} 

\graphicspath{{./Images/}{./}}

\hypersetup{
  hidelinks,
  pdfauthor={Yi Liu, TingFeng Hui, Wei Zhang, Li Sun, Ningxin Su, Jian Wang, Sen Su},
  pdftitle={EnvSimBench: A Benchmark for Evaluating and Improving LLM-Based Environment Simulation}
}

\iclrfinalcopy

\definecolor{dsblue}{HTML}{35e7df}
\definecolor{gptorange}{HTML}{2c97ff}
\definecolor{geminiteal}{HTML}{028104}
\definecolor{claudepurple}{HTML}{0101ff}
\definecolor{qwenred}{HTML}{dc0d0d}
\definecolor{glmgold}{HTML}{fddc32}
\definecolor{minimaxgreen}{HTML}{2ad92d}
\definecolor{balance2col}{HTML}{000000}

\definecolor{lightgrey}{HTML}{A5BAD1}
\definecolor{lightcyan}{HTML}{d5e5c9}
\definecolor{lightyellow}{HTML}{eea78b}
\definecolor{lightpink}{HTML}{e3c7d5}

\title{EnvSimBench: A Benchmark for Evaluating and Improving LLM-Based Environment Simulation}

\author{%
  \parbox{0.97\linewidth}{\centering
    Yi Liu\textsuperscript{1} \quad
    TingFeng Hui\textsuperscript{1} \quad
    Wei Zhang\textsuperscript{1} \quad
    Li Sun\textsuperscript{1} \quad
    Ningxin Su\textsuperscript{2} \quad
    Jian Wang\textsuperscript{3} \quad
    Sen Su\textsuperscript{1} \\[0.3em]
    \mdseries
    \textsuperscript{1}\mbox{Beijing University of Posts and Telecommunications} \quad
    \textsuperscript{2}\mbox{The Hong Kong University of Science and Technology} \quad
    \textsuperscript{3}\mbox{Chongqing University} \\[0.5em]
    \faEnvelope\ \texttt{\mdseries louie@bupt.edu.cn}
  }
}

\begin{document}  

\maketitle

\begin{abstract} 
LLM-based environment simulators offer a flexible substrate for agent training, but their usefulness depends on whether generated feedback and state updates faithfully reflect agent actions. 
In practice, LLM-simulated environments suffer from hallucinations, logical inconsistencies, and silent state drift—failures that corrupt agent reward signals and compound the construction costs that the paradigm was designed to eliminate.
To address this gap, we propose EnvSimBench with four contributions:
(1) We provide the first formal definition and operationalization of \textbf{Environment Simulation Ability (EnvSim Ability)} as a quantifiable research objective.
(2) We construct EnvSimBench, a rigorous benchmark covering 400 samples across 167 diverse environments, equipped with verifiable labels and fine-grained difficulty stratification along three axes. 
(3) Evaluations of seven frontier language models reveal a pronounced decline in Config Match on state-changing operations relative to state-preserving operations, which we term the state-change cliff. This finding identifies a limitation in predicting the state updates required by the evaluated environments.
(4) We design a constraint-driven simulation pipeline that substantially reduces hallucination, boosts environment synthesis yield by 6.8\%, and cuts costs by over 90\%.
Overall, EnvSimBench serves as both a diagnostic framework and a practical optimization path for reliable LLM-based environment simulation, establishing a solid foundation for scalable agent training. Code and data are available at \url{https://github.com/cookieApril/EnvSimBench}.
\end{abstract}

\section{Introduction}
\label{sec:intro}
Large language models (LLMs) are increasingly required to serve as agents across diverse real-world scenarios~\citep{luo2025largelanguagemodelagent,yao2025taubench,qian2025userbenchinteractivegymenvironment}. Recent research efforts~\citep{patil2025bfcl,yao2025taubench,lu2025toolsandboxstatefulconversationalinteractive} have constructed stateful, tool-interactive sandbox environments, delivering improved controllability and stability.

Training autonomous agents at scale requires interactive environments that feature rich diversity, high simulation fidelity, and low deployment costs~\citep{huang2025scaling,deepseekai2025deepseekv32pushingfrontieropen,froger2025arescalingagentenvironments}. However, manually designed environments suffer from prohibitive construction costs, poor extensibility, and inherent limitations in scenario coverage, making them ill-suited for large-scale agent training. 
A compelling recent direction addresses this challenge by replacing manually crafted, executable environments\citep{jimenez2024swebench, zhou2024webarena,shridhar2021alfworld} with LLM-simulated counterparts, where a language model generates the feedback for each agent action~\citep{li2025simulatingenvironmentsreasoningmodels,
song2026envscalerscalingtoolinteractiveenvironments}. This paradigm dramatically lowers the barrier to environment creation and promises seamless scaling across domains. However, it rests on a foundational assumption that has received surprisingly little scrutiny: LLMs can accurately simulate environmental feedback with sufficient fidelity to be trusted as training substrates.

If this assumption fails, agents trained in hallucinated environments optimize against corrupted reward signals, and synthesis pipelines incur compounding failure costs that erode the very cost advantage the paradigm promises, but simulation fidelity has never been systematically characterized.

In practice, when models attempt to act as simulators, their performance is undermined by specific vulnerabilities. 
Three concrete failure modes underlie this difficulty: hallucination~\citep{ji2023hallucination,patil2023gorilla,zhang2025sirenssongaiocean}, whereby a model fabricates plausible but incorrect state transitions; logical inconsistency~\citep{ruan2024identifyingriskslmagents,elazar-etal-2021-measuring}, whereby inter-field constraints within a single response are violated; and state drift~\citep{lu2025toolsandboxstatefulconversationalinteractive,prabhakar2025apigenmtagenticpipelinemultiturn,yao2025taubench}, whereby the absence of persistent memory causes earlier state changes to be silently lost. These failures are not incidental engineering deficiencies but reflect a fundamental tension between the generative nature of language models and the deterministic, compositional logic of executable environments. Without a framework to precisely measure these failures, they remain invisible to practitioners and unaddressable by researchers.

To address this gap and establish a rigorous foundation for studying simulation fidelity, we introduce the concept of Environment Simulation Ability (EnvSim Ability) to denote a model's capacity to accurately predict the state transition and observational feedback induced by an agent action, given the current environment state and the action's implementation logic. Without a precise, operationalizable definition, neither systematic evaluation nor principled improvement of this capability is possible. Executable environments provide reference transitions against which simulator predictions can be checked. Their role in EnvSimBench is to make transition errors observable and reproducible. This evaluation quantifies how reliably a model reproduces specified environment behavior; transfer to environments with implicit rules or unavailable execution references requires separate validation.

To address three failure modes and make EnvSim Ability concretely measurable, we construct EnvSimBench. EnvSimBench makes the recorded configuration explicit for independently verifiable transition prediction. In a history-conditioned simulation setting, the current state is not directly supplied, giving a partially observable formulation~\citep{oliehoek2016concise} in which the simulator must infer the state from preceding interactions. EnvSimBench separates transition prediction from history-based state estimation. We provide the recorded pre-execution configuration and tool implementation for each sample. This explicit-state protocol isolates errors in predicting the next observation and configuration from errors accumulated while reconstructing earlier states. It enables a controlled diagnosis of transition fidelity, a component of reliable multi-turn simulation. This reframing is the architectural foundation of EnvSimBench.

EnvSimBench makes four contributions, linked by a deliberate logical chain from conceptualization to diagnosis to remedy:

\textbf{(I)} \textbf{Formalizing EnvSim Ability.}
We provide the first formal definition and operationalization of EnvSim Ability as a quantifiable research objective, establishing it as a distinct capability from related skills and providing a conceptual foundation for rigorous evaluation and improvement.

\textbf{(II)} \textbf{A rigorous benchmark.}
Grounded in the POMDP-to-MDP reframing, we construct a benchmark of 400 samples across 167 diverse tool-interactive environments, with verifiable programmatic labels and fine-grained difficulty stratification along three stratification dimensions: action outcome, state-change complexity, and argument cardinality.

\textbf{(III)} \textbf{A diagnostic finding: the state-change cliff.}
Systematic evaluation reveals a universal failure pattern: all models achieve near-perfect accuracy on state-preserving operations yet collapse catastrophically when multiple state variables must update simultaneously. We term this threshold the \emph{state-change cliff}. This gap is orthogonal to model scale and general reasoning ability. More critically, models that generate superficially correct feedback can simultaneously produce incorrect state transitions, silently corrupting training signals with no observable divergence for the agent.

\textbf{(IV)} \textbf{A constraint-driven remedy.}
Guided by these findings, we propose a constraint-driven simulation paradigm that makes environment schemas and transition logic explicit at each step. A simulation model trained under this paradigm surpasses all evaluated frontier LLMs on configuration match, boosts synthesis yield by 6.8\%, and cuts costs by over 90\%. This demonstrates that targeted specialization is a cost-efficient path to reliable environment simulation.

\begin{figure}[t]
    \centering
    \vspace{-15pt}
    \includegraphics[width=\linewidth]{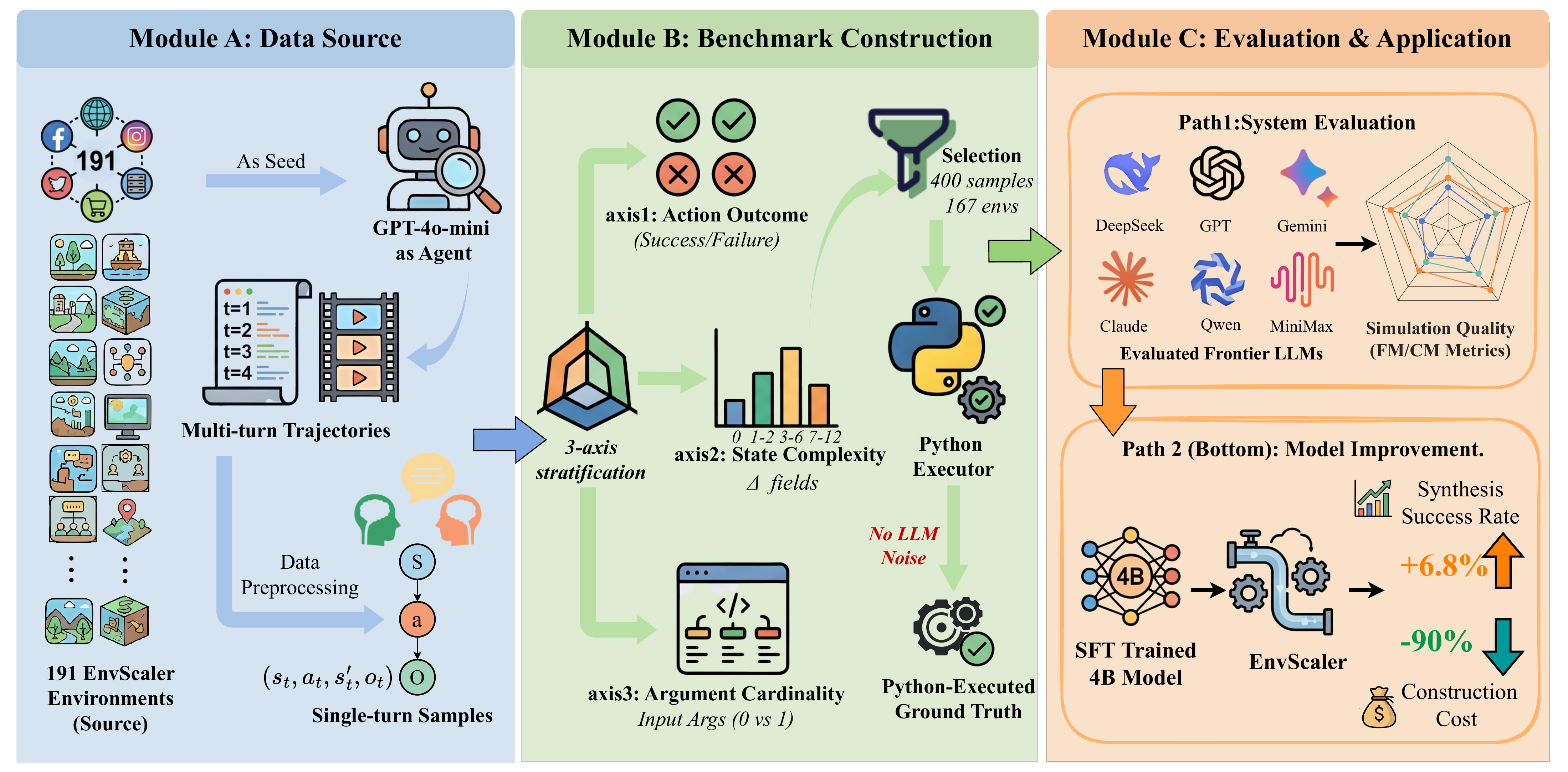}
    \caption{%
        \textbf{Overview of EnvSimBench.}
        \textbf{Module A:} EnvScaler~\citep{song2026envscalerscalingtoolinteractiveenvironments} environments serve as seed data; a GPT-4o-mini agent collects multi-turn execution trajectories, preprocessed into self-contained single-turn state prediction samples $(s_t, a_t, s'_t, o_t)$. Each step is independently verifiable against a programmatic label, decoupling simulation fidelity from state tracking and making EnvSim Ability objectively measurable.
        \textbf{Module B:} Samples undergo three-axis stratification (action outcome, state-change complexity, argument cardinality) and executor-based labeling, yielding 400 benchmark samples across 167 environments. The three axes allow failures to be precisely localized rather than collapsed into a single score.
        \textbf{Module C:} \emph{Path~1} evaluates seven frontier LLMs under identical conditions. \emph{Path~2} trains a specialized 4B simulation model that, when integrated into EnvScaler, improves synthesis yield by 6.8\% while cutting costs by over 90\%.
    }
    \label{fig:framework}
\end{figure}

\section{Related Work}
\label{sec:related}

\paragraph{LLM environment simulation and synthesis.}
Simia~\citep{li2025simulatingenvironmentsreasoningmodels} demonstrated that reasoning models can generate plausible tool feedback for agent training, establishing LLM-simulated tool-interactive environments as a practical substrate for scalable agent training. EnvScaler~\citep{song2026envscalerscalingtoolinteractiveenvironments} extended this direction by demonstrating a fully automated synthesis of tool-interactive environments, substantially reducing construction costs. However, neither work formally defines simulation fidelity as a measurable capability, provides cross-model comparisons of simulation quality, or offers principled evaluation metrics. EnvSimBench addresses all three gaps simultaneously. Relatedly, interactive fiction environments~\citep{hausknecht2020interactivefictiongamescolossal} and code-execution benchmarks~\citep{jimenez2024swebench} use executable environments to evaluate agent behavior; our work inverts this relationship, treating the simulator itself rather than the agent as the object of evaluation.

\paragraph{Agent benchmarks and tool use.}
Building on a long line of prior work on language agents~\citep{yao2023webshopscalablerealworldweb,zhou2024webarenarealisticwebenvironment,jimenez2024swebenchlanguagemodelsresolve,liu2025agentbenchevaluatingllmsagents,ruan2024identifyingriskslmagents,deng2023mind2webgeneralistagentweb,zeng2023agenttuningenablinggeneralizedagent}, LLM tool utilization~\citep{patil2025bfcl, ICLR2024_28e50ee5,huang2024metatool}, and task-oriented dialogue systems $\tau$-bench~\citep{yao2025taubench} and $\tau^2$-bench~\citep{barres2025tau2benchevaluatingconversationalagents} established multi-turn, tool-using evaluation paradigms and introduced the MDP/POMDP distinction that we adopt as the architectural foundation of our diagnostic framework. ToolSandbox~\citep{lu2025toolsandboxstatefulconversationalinteractive} evaluates stateful tool use with fine-grained state-transition feedback, providing a complementary perspective in which the agent is evaluated against a fixed executable environment; our work evaluates the simulator rather than the agent, and asks whether a language model can \emph{replace} such executors reliably. 
APIGen-MT~\citep{prabhakar2026apigenmt} generates multi-turn training data via LLM simulation; our pipeline shares structural similarities but replaces LLM-generated labels with programmatically verified ground truth, directly addressing the circular validation problem that motivates our work.

\paragraph{Hallucination and simulation fidelity.}
Hallucination in language model generation has been extensively studied in open-ended settings~\citep{ji2023hallucination, zhang2025sirenssongaiocean}. Gorilla~\citep{patil2023gorilla} demonstrated that language models frequently produce incorrect tool arguments even when the correct API schema is provided. This finding also highlights the importance of our research. Prior work on language model program execution~\citep{lu2025toolsandboxstatefulconversationalinteractive} shows that prediction accuracy degrades as the number of interdependent state updates increases, underscoring the cumulative effect of state updates across multiple execution rounds. EnvSimBench prevented the cascading dependency identified in this article by using the POMDP-to-MDP pattern. 
 
\section{Problem Formulation}
\label{sec:problem}
 
\paragraph{Tool interactive environments.}
Following EnvScaler, we model a tool-interactive environment as $\mathcal{E}=(\mathcal{S},\mathcal{A},\mathcal{T},\mathcal{O})$, where $\mathcal{S}$ is the configuration space and $\mathcal{A}$ is the action space. For fixed runtime inputs $\xi$, the transition and observation functions are deterministic: \[   s'_t=\mathcal{T}(s_t,a_t;\xi_t),   \qquad   o_t=\mathcal{O}(s_t,a_t;\xi_t)\in\Sigma^*. \] Here $\xi_t$ denotes runtime inputs such as clock values used during execution. For example, a mutual fund management system stores securities, portfolio holdings, and transaction logs, while a mobile authentication system stores OTP records, session states, and verification timestamps. Reference outputs $(o_t,s'_t)$ are obtained by executing the tool implementation. Runtime inputs need not be recoverable from the stored configuration; their implications for exact matching are discussed in Appendix~\ref{app:timestamp_analysis}.
 
\paragraph{State prediction task.}
Given the pre-execution configuration $s_t$, a tool call $a_t$, and its implementation context $c_t=\texttt{code}(a_t)$, the model predicts a feedback string $\hat{o}_t$ and a list of state-change operations $\widehat{\Delta}_t$. Each operation specifies an addition, modification, or deletion at a configuration path. Let $\Delta_t=\operatorname{Diff}(s_t,s'_t)$ denote the reference changes recorded by the benchmark's configuration-difference procedure, and let $k_t=|\Delta_t|$ denote the state-change count. The predicted list $\widehat{\Delta}_t$ may contain a different number of operations. The predicted after-state is reconstructed as
\begin{equation}
  (\hat{o}_t,\widehat{\Delta}_t)=f_\theta(s_t,a_t,c_t),
  \qquad
  \hat{s}'_t=\operatorname{Apply}(s_t,\widehat{\Delta}_t).
  \label{eq:transition_prediction}
\end{equation}
Feedback Match (FM) tests exact equality between $\hat{o}_t$ and $o_t$, while Config Match (CM) tests equality between the reconstructed $\hat{s}'_t$ and $s'_t$. Both metrics use executor-produced reference outputs.

\paragraph{POMDP vs.\ MDP formulation.}
In history-conditioned simulation, the current configuration is not supplied explicitly. Let $h_t=(a_1,o_1,\ldots,a_{t-1},o_{t-1})$ denote the preceding interaction history. A belief state is a distribution $b_t(s)=\Pr(S_t=s\mid h_t)$ over possible current configurations. A history-conditioned simulator predicts the next output from $(h_t,a_t)$, combining state reconstruction, transition prediction, and observation generation. When predicted outputs are reused in later interactions, earlier errors can affect subsequent predictions.

\paragraph{Constraint-driven MDP formulation.}
EnvSimBench supplies the recorded pre-execution configuration for every prediction, together with the current tool call and its implementation context, as in Equation~\ref{eq:transition_prediction}. Each sample is scored against its own executor-produced reference output. Consequently, an incorrect prediction on one sample is not propagated as the input state of another. This protocol isolates single-step transition fidelity from history reconstruction while retaining errors in code interpretation, feedback generation, and state updates as measurable outcomes.
 
\begin{figure}[t]
    \centering
    \includegraphics[width=0.92\linewidth]{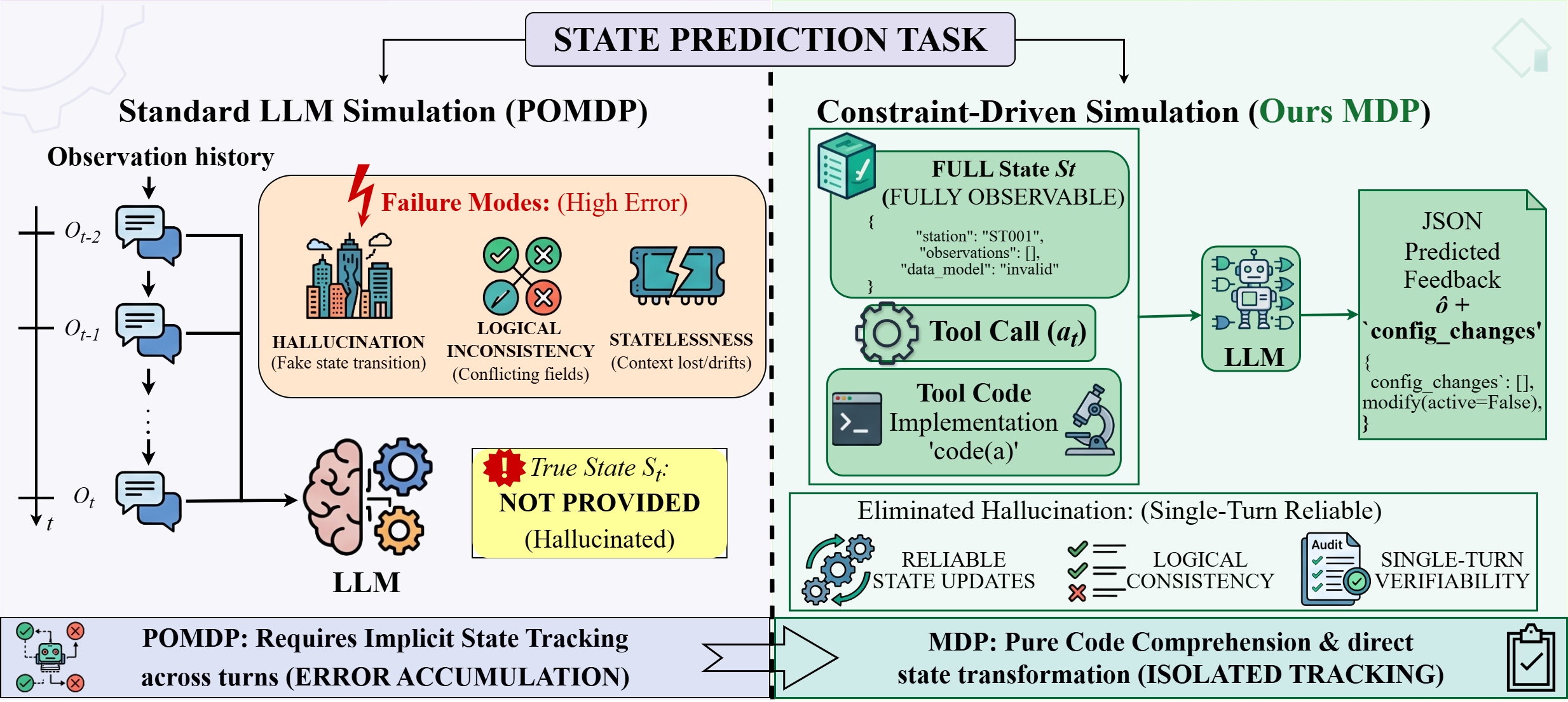}
    \caption{%
        \textbf{POMDP vs.\ MDP formulation.}
        \emph{Left}: Standard simulation; the LLM infers state from conversation history, causing state drift and hallucination.
        \emph{Right}: Constraint-driven simulation; the full before-config $s_t$, tool call $a_t$, and implementation \texttt{code(a)} are provided explicitly, making each step independently verifiable.
    }
    \label{fig:pomdp_vs_mdp}
\end{figure}
 
\section{EnvSimBench and Constraint-Driven Simulation Paradigm}
\label{sec:bench}

\subsection{Benchmark Construction}

\paragraph{From trajectories to single-turn samples.}
The constraint-driven formulation (Section~\ref{sec:problem}) motivates self-contained samples: each supplies $(s_t,a_t,c_t)$ and requires $(\hat{o}_t,\widehat{\Delta}_t)$, from which the after-state is reconstructed. Each prediction is evaluated against the recorded reference transition, without using another prediction as its input state.

As illustrated in Figure~\ref{fig:benchmark_construction}, we collect execution trajectories $\tau=((a_t,o_t,s_t,s'_t))_{t=1}^{T}$ from 191 tool-interactive seed environments. We discard the initialization-only record because it does not contain a tool call and a corresponding before/after configuration pair. The configuration produced by initialization serves as the before-state of the first retained tool call. Each retained sample has an executor-produced reference output $(o_t,s'_t)$ and is evaluated separately.

\paragraph{Three-axis difficulty stratification.}
To enable fine-grained diagnosis rather than a single aggregate score, we stratify the sample pool along three dimensions, each targeting a source of simulation difficulty. Figure~\ref{fig:framework} provides an overview of EnvSimBench, and a detailed construction pipeline is described in Appendix~\ref{app:construction}.

\textit{Axis~1 (Action Outcome).}
Samples are first partitioned by whether the action $a_t$ succeeds or fails under $\mathcal{T}$. The \emph{Failure} group ($|\Delta(s_t, s'_t)| = 0$ with $\mathcal{O}$ returning an error signal) tests whether a model correctly withholds state changes when the transition guard is not satisfied.

\textit{Axis~2 (State-Change Complexity).}
Within the success group, samples are stratified by $|\Delta(s_t,s'_t)|$, the number of changes recorded by the configuration-difference procedure, into four tiers: \emph{No-Change} ($|\Delta|=0$, 80 samples), \emph{Simple} ($|\Delta|\in\{1,2\}$, 50 samples), \emph{Medium} ($|\Delta|\in\{3,\ldots,6\}$, 200 samples), and \emph{Difficult} ($|\Delta|\in\{7,\ldots,12\}$, 50 samples). The largest allocation is assigned to Medium samples to support analysis of multi-field updates. State-change count is a stratification variable; update structure can also affect prediction difficulty (Appendix~\ref{app:difficult_pattern}).

\textit{Axis~3 (Input Argument Cardinality).}
For Failure and No-Change samples, a third axis further partitions by the number of arguments supplied to $a_t$ (zero vs.\ one or more). This controls for a qualitatively simpler sub-case: when no arguments are provided, the model need not reason about argument-dependent branching in the transition logic, making correct prediction structurally easier regardless of state-change complexity.

\paragraph{Diversity rule.}
Within each stratum, samples are selected to maximize the number of distinct environments $|\mathcal{E}_{\text{sub}}|$ represented. This \emph{diversity rule} prevents the benchmark from concentrating in a small number of high-trajectory environments and ensures that evaluation generalizes across the full breadth of domain types present in the source collection. The final benchmark comprises 400 samples drawn from 167 distinct environments (Table~\ref{tab:sample_distribution}). All ground-truth labels are produced by the deterministic external executor, making the benchmark entirely LLM-free in its construction.

\begin{table}[htbp]
\centering
\small
\caption{Benchmark sample distribution.}
\label{tab:sample_distribution}
\begin{tabular}{llcc}
\toprule
\textbf{Group} & \textbf{Subgroup} & \textbf{Samples} & \textbf{Constraint} \\
\midrule
Failure   & $\mathcal{O}$ returns error, $|\Delta|=0$  & 20 &  \\
No-Change & $|\Delta|=0$, action succeeds              & 80  & 40 per argument cardinality \\
Simple    & $|\Delta| \in \{1,2\}$                    & 50  & 25 per $|\Delta|$ value \\
Medium    & $|\Delta| \in \{3,\ldots,6\}$             & 200 & 50 per $|\Delta|$ value \\
Difficult & $|\Delta| \in \{7,\ldots,12\}$            & 50  & Distributed across $|\Delta|$ \\
\bottomrule
\end{tabular}
\end{table}

\paragraph{Evaluation metrics.}
For each sample $i$, we evaluate feedback and configuration correctness separately:
\begin{equation}
  \mathrm{FM}_i=\mathbf{1}[\hat{o}_i=o_i],
  \qquad
  \mathrm{CM}_i=\mathbf{1}[\operatorname{Apply}(s_i,\widehat{\Delta}_i)=s'_i].
  \label{eq:match_metrics}
\end{equation}
FM requires exact agreement with the reference feedback string. CM requires the reconstructed configuration to match the reference after-state. The two metrics measure complementary aspects of a simulated transition: matching the feedback does not guarantee a correct state update, and a correct state update does not guarantee matching feedback. For a group $G$, we report the sample mean of each binary metric.

\subsection{Constraint-Driven Simulation Paradigm}

\paragraph{MDP prompt structure.}
Each evaluation prompt supplies $(s_t,a_t,c_t)$ in a structured format. The response contains $\hat{o}_t$ and a list $\widehat{\Delta}_t=(\delta_{t,1},\ldots,\delta_{t,m_t})$ of typed state-change operations. Each operation identifies its type, target path, and a value when required. The evaluator reconstructs $\hat{s}'_t=\operatorname{Apply}(s_t,\widehat{\Delta}_t)$ and compares it with $s'_t$ to compute CM.

\paragraph{Reasoning enhancement.}
Training data is augmented with explicit intermediate derivations, formatted
as structured reasoning traces that walk through the transition logic of
$\texttt{code}(a_t)$ step by step before producing $(\hat{o}_t,
\widehat{\Delta})$. These traces teach the supervised model to internalize
code-execution as a sequential inference process rather than a pattern-matching
lookup.

\section{Experimental Results and Analysis}
\label{sec:experiments}

\subsection{Evaluation of Frontier LLMs}
\label{sec:frontier}

\paragraph{Setup.}
We evaluate seven frontier language models: DeepSeek-V3.2, Qwen3.5-397B-A17B, GPT-5.4, Gemini-3.1-Pro-Preview, Claude-Sonnet-4.6, MiniMax-M2.7, and GLM-5, all in non-thinking mode via their respective inference APIs, with identical prompts instantiating the MDP triple $(s_t, a_t, \texttt{code}(a_t))$. Ground-truth labels are produced entirely by the deterministic external executor. Full evaluation details are provided in Appendix~\ref{app:eval_setup}. Supplementary reasoning-mode results are reported in Appendix~\ref{app:reasoning_eval}.
 
\begin{table}[htbp]
\centering
\small
\setlength{\tabcolsep}{4pt}
\caption{Results on Failure and No-Change samples.
  \textbf{Bold}: column-wise maximum;
  \textcolor{red}{red}: column-wise minimum.}
\label{tab:fail_nochange}
\begin{tabular}{l cc cc cc cc}
\toprule
& \multicolumn{2}{c}{\textbf{Fail}}
& \multicolumn{2}{c}{\textbf{No-Chg(0)}}
& \multicolumn{2}{c}{\textbf{No-Chg(1)}}
& \multicolumn{2}{c}{\textbf{Avg}} \\
\cmidrule(lr){2-3}\cmidrule(lr){4-5}\cmidrule(lr){6-7}\cmidrule(lr){8-9}
\textbf{Model} & FM & CM & FM & CM & FM & CM & FM & CM \\
\midrule
DeepSeek-V3.2
  & 85\%                  & 100\%
  & \textbf{85\%}         & \textbf{100\%}
  & \textbf{82.5\%}       & 100\%
  & \textbf{84\%}         & \textbf{100\%} \\
Qwen3.5-397B-A17B
  & 90\%                  & 100\%
  & \textbf{85\%}         & \textbf{100\%}
  & 77.5\%                & 100\%
  & 83\%                  & \textbf{100\%} \\
GPT-5.4
  & 90\%                        & 100\%
  & \textcolor{red}{65\%}       & \textbf{100\%}
  & \textcolor{red}{67.5\%}     & 100\%
  & \textcolor{red}{71\%}       & \textbf{100\%} \\
Gemini-3.1-Pro-Preview
  & \textbf{95\%}         & 100\%
  & 77.5\%                & \textbf{100\%}
  & 70\%                  & 100\%
  & 78\%                  & \textbf{100\%} \\
Claude-Sonnet-4.6
  & \textcolor{red}{15\%} & 100\%
  & 80\%                  & \textcolor{red}{97.5\%}
  & 77.5\%                & 100\%
  & 66\%                  & \textcolor{red}{99\%} \\
MiniMax-M2.7
  & 80\%                        & 100\%
  & 67.5\%                      & \textcolor{red}{97.5\%}
  & \textcolor{red}{62.5\%}     & 100\%
  & \textcolor{red}{68\%}       & \textcolor{red}{99\%} \\
GLM-5
  & 90\%   & 100\%
  & 77.5\% & \textbf{100\%}
  & 70\%   & 100\%
  & 77\%   & \textbf{100\%} \\
\bottomrule
\end{tabular}
\end{table}
\paragraph{Failure and No-Change groups.}
Table~\ref{tab:fail_nochange} shows that CM is near-perfect (97--100\%) across all state-preserving operations, confirming that every model correctly withholds predicted state changes when none are warranted. FM exhibits wider variation. Notably, Claude-Sonnet-4.6 achieves only 15\% FM on the Failure group despite 100\% CM. This divergence reveals a systematic output-format mismatch: the model omits the enclosing structured response schema and returns only the inner message string. Concretely, whereas the ground-truth observation $o_t$ takes the form of a key-value structured response $\langle \textit{status}: \text{false},\, \textit{msg}: \varepsilon \rangle$, the model outputs only $\varepsilon$ directly. MiniMax-M2.7 exhibits the same unwrapping behavior on success-type responses. Because this mismatch is an artifact of a formatting convention rather than a reasoning failure, CM remains unaffected. We adopt CM as the primary cross-model reasoning metric throughout.

\begin{table}[htbp]
\centering
\small
\setlength{\tabcolsep}{4pt}
\caption{Averaged results on state-changing groups. CM degrades sharply with $|\Delta|$.
  \textbf{Bold}: column-wise maximum;
  \textcolor{red}{red}: column-wise minimum.}
\label{tab:statechange_summary}
\begin{tabular}{l cc cc cc}
\toprule
& \multicolumn{2}{c}{\textbf{Simple (avg)}}
& \multicolumn{2}{c}{\textbf{Medium (avg)}}
& \multicolumn{2}{c}{\textbf{Difficult (avg)}} \\
\cmidrule(lr){2-3}\cmidrule(lr){4-5}\cmidrule(lr){6-7}
\textbf{Model} & FM & CM & FM & CM & FM & CM \\
\midrule
DeepSeek-V3.2
  & 80\%                        & \textcolor{red}{22\%}
  & 76.5\%                      & \textcolor{red}{8.5\%}
  & 26\%                        & \textcolor{red}{4\%} \\
Qwen3.5-397B-A17B
  & 84\%   & 46\%
  & 65.0\% & 17.0\%
  & 42\%   & 24\% \\
GPT-5.4
  & 80\%          & 40\%
  & 75.0\%        & \textbf{17.5\%}
  & \textbf{74\%} & 26\% \\
Gemini-3.1-Pro-Preview
  & \textbf{96\%} & 48\%
  & 68.0\%        & \textbf{17.5\%}
  & 68\%          & 18\% \\
Claude-Sonnet-4.6
  & \textcolor{red}{10\%}   & 40\%
  & \textcolor{red}{12.5\%} & 12.0\%
  & \textcolor{red}{12\%}   & 16\% \\
MiniMax-M2.7
  & 32\%   & \textbf{50\%}
  & 22.0\% & 16.0\%
  & \textcolor{red}{8\%}  & 22\% \\
GLM-5
  & 94\%            & 44\%
  & \textbf{81.5\%} & 14.0\%
  & 70\%            & \textbf{28\%} \\
\bottomrule
\end{tabular}
\end{table}

\paragraph{State-changing groups (Simple / Medium / Difficult).}
Table~\ref{tab:statechange_summary} summarizes CM and FM across groups requiring actual state updates. CM degrades sharply with $|\Delta|$: most models fall below 20\% on Medium and approach zero on Difficult, while FM remains substantially higher throughout. This persistent FM--CM gap reveals that models can generate surface-plausible response strings for an action without faithfully tracing the induced state transition $\mathcal{T}(s_t, a_t)$. Detailed per-$|\Delta|$ breakdowns appear in Appendix~\ref{app:detailed_results}.


\begin{table}[htbp]
\centering
\small
\setlength{\tabcolsep}{4pt}
\caption{Overall results. CM is robust to output-format variation; FM is depressed
for models with format-unwrapping behavior (Claude-Sonnet-4.6, MiniMax-M2.7).
  \textbf{Bold}: column-wise maximum;
  \textcolor{red}{red}: column-wise minimum.}
\label{tab:overall_results}
\begin{tabular}{l cc cc cc}
\toprule
& \multicolumn{2}{c}{\textbf{Fail+No-Chg}}
& \multicolumn{2}{c}{\textbf{State-Change}}
& \multicolumn{2}{c}{\textbf{Overall}} \\
\cmidrule(lr){2-3}\cmidrule(lr){4-5}\cmidrule(lr){6-7}
\textbf{Model} & FM & CM & FM & CM & FM & CM \\
\midrule
DeepSeek-V3.2
  & \textbf{84\%}         & \textbf{100\%}
  & 68.7\%                & \textcolor{red}{10.0\%}
  & 72.5\%                & \textcolor{red}{32.5\%} \\
Qwen3.5-397B-A17B
  & 83\%   & \textbf{100\%}
  & 64.3\% & \textbf{23.0\%}
  & 69.0\% & \textbf{42.3\%} \\
GPT-5.4
  & 71\%   & \textbf{100\%}
  & 75.7\% & 22.7\%
  & 74.5\% & 42.0\% \\
Gemini-3.1-Pro-Preview
  & 78\%   & \textbf{100\%}
  & 72.7\% & 22.7\%
  & 74.0\% & 42.0\% \\
Claude-Sonnet-4.6
  & \textcolor{red}{66\%} & \textcolor{red}{99\%}
  & \textcolor{red}{12.0\%} & 17.3\%
  & \textcolor{red}{25.5\%} & 37.8\% \\
MiniMax-M2.7
  & 68\%   & \textcolor{red}{99\%}
  & 21.3\% & 22.7\%
  & 33.0\% & 41.8\% \\
GLM-5
  & 77\%            & \textbf{100\%}
  & \textbf{81.7\%} & 21.3\%
  & \textbf{80.5\%} & 41.0\% \\
\bottomrule
\end{tabular}
\end{table}

\noindent\textbf{Finding 1: Substantial capability gaps exist across frontier LLMs.}
Overall CM ranges from 32.5\% (DeepSeek-V3.2) to 42.3\% (Qwen3.5-397B-A17B), a nearly 10~pp spread that originates \emph{entirely} from state-changing samples, since all models achieve $\geq$99\% CM on state-preserving operations. On state-changing samples, DeepSeek-V3.2 achieves only 10.0\% CM while the best frontier models reach 22--23\%. This gap is not explained by model scale: DeepSeek-V3.2 scores competitively on standard reasoning benchmarks yet exhibits the largest CM deficit here, suggesting that \emph{EnvSim Ability} is a capability orthogonal to general reasoning proficiency (strong reasoning does not imply strong simulation).

\noindent\textbf{Finding 2 (State-change cliff): All models collapse in CM when
$|\Delta| \geq 3$.}
Per-$|\Delta|$ results in Appendix~\ref{app:detailed_results} reveal a universal threshold. At $|\Delta|=1$, CM spans 36--72\%; by $|\Delta|=5$, every model falls to $\leq$4\%. Across all samples with $|\Delta| \geq 5$, CM rates converge near zero: DeepSeek-V3.2 2.7\%, Claude-Sonnet-4.6 6.0\%, Gemini-3.1-Pro-Preview 9.3\%, MiniMax-M2.7 10.7\%, GPT-5.4 12.0\%, GLM-5 11.3\%. This threshold marks a \emph{qualitative} transition rather than a gradual decline. Below it ($|\Delta| \leq 4$), models at least partially track state changes: getting some fields right while missing others. Above it ($|\Delta| \geq 5$), predictions collapse: many models output $\widehat{\Delta} = \emptyset$, reporting no state change at all, even though the corresponding real action actually needs to modify a large number of state fields.
A partial CM recovery visible at $|\Delta|=7$ for several models is not evidence of improved capability but reflects a structural artifact: high-$|\Delta|$ samples that models answer correctly are dominated by \emph{bulk-uniform} operations whose changes instantiate a single repeating template, a qualitatively easier task than the heterogeneous multi-field updates that cause the cliff (see Appendix~\ref{app:difficult_pattern}).

\noindent\textbf{Finding 3: FM and CM capture complementary error types.}
\emph{Format mismatch} can cause FM failure despite a correct state prediction. For example, Claude-Sonnet-4.6 has 15\% FM and 100\% CM on Failure samples, but its FM is 80\% and 77.5\% on the two No-Change subgroups. Format sensitivity therefore does not imply uniformly low FM across state-preserving operations.

\emph{State-tracking failure} is universal and $|\Delta|$-dependent, manifesting in three sub-patterns: \textbf{(i)} models systematically neglect auxiliary state updates induced as side effects of the primary action; \textbf{(ii)} fields whose values are determined only at execution time (such as clock-generated timestamps) cannot be correctly predicted without actually running the transition logic; and \textbf{(iii)}, most consequentially, a model may produce $\hat{o}_t = o_t$ while simultaneously computing $\hat{s}'_t \neq s'_t$, giving the agent no divergence signal whatsoever. On $|\Delta| \geq 3$ samples where CM fails, 50--64\% of failures also have FM\,$= 1$ for DeepSeek-V3.2, GLM-5, GPT-5.4, and Gemini-3.1-Pro-Preview. This means the environment state is silently corrupted while the agent's reward signal remains intact. Quantitative breakdowns and cross-model case studies for all three sub-patterns are provided in Appendix~\ref{app:failure_analysis}.

\subsection{Evaluation of Small Models: Before and After SFT}
\label{sec:sft}

\paragraph{Setup.}
We fine-tune Qwen3-4B-Base via full-parameter SFT on 2$\times$A800 (80\, GB) GPUs, evaluating on an \texttt{env\_id} holdout set that excludes all 167 benchmark environments from training. A pre-SFT scaling baseline confirms that specialization is required beyond scale alone (Appendix~\ref{app:scaling}). Full hyperparameter details and training method comparisons are provided in Appendix~\ref{app:eval_setup}.
Full-parameter SFT substantially improves FM ($27\% \to 72.8\%$, $2.7\times$) and corrects output-schema alignment; the state-change cliff persists at the 5K-sample scale, confirming it as a fundamental capability gap. Detailed per-group results and a comparison of training strategies are reported in Appendix~\ref{app:training_method}.

\begin{figure}[tbp]
  \centering
  \includegraphics[width=0.95\linewidth]{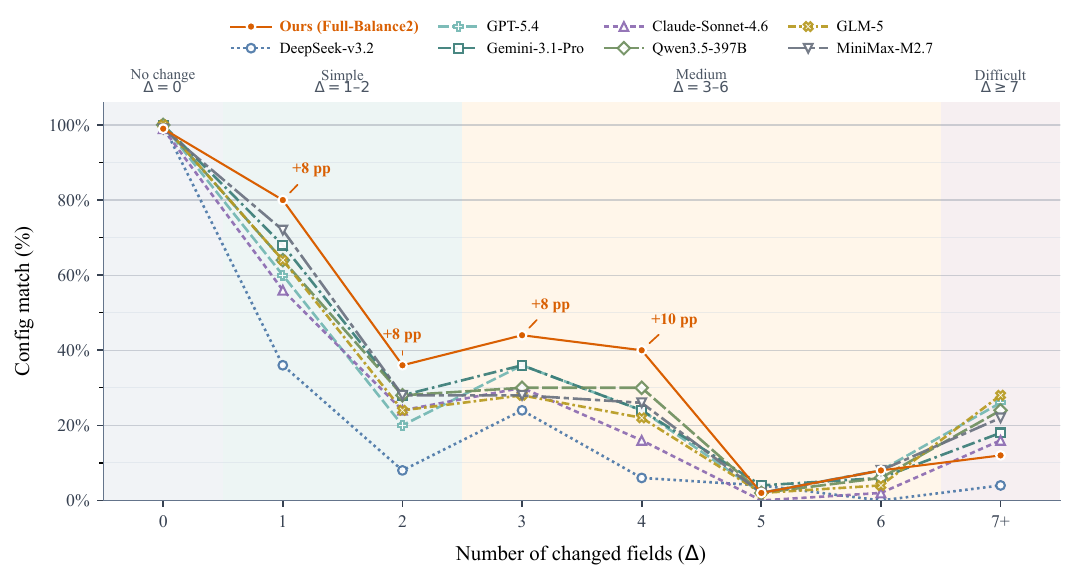}
  \caption{\textbf{Config Match by state-change count on the original benchmark.}
  Full-Balance2 (orange) is compared with the seven non-thinking frontier baselines.
  At $|\Delta|\in\{1,2,3,4\}$, Full-Balance2 exceeds the strongest frontier baseline at each level by up to 10~pp.
  The final point pools samples with $|\Delta|=7$--12; the trend is non-monotonic.}
  \label{fig:cliff_and_sft}
\end{figure}

\noindent\textbf{Finding 4: Reasoning-augmented traces do not consistently improve CM at 5K samples.}
As shown in Table~\ref{tab:sft_method}, full SFT with structured reasoning traces achieves 30.3\% CM, compared with 35.0\% without structured reasoning, a 4.7~pp regression. Traces hurt on Simple samples (20\% vs.\ 34\%), likely adding distributional noise where transition logic is short, while providing a modest gain on Difficult samples (10\% vs.\ 6\%). We speculate that consistent benefits require either substantially more data or difficulty-conditioned routing that applies traces selectively to complex samples.

\noindent\textbf{Finding 5 (Data composition): Composition governs generalization more strongly than volume.}
We train \textbf{Balance2} by mirroring the empirical difficulty distribution of $|\Delta|$ across source environments: 1{,}000 failure $+$ 1{,}000 no-change $+$ 2{,}000 simple-change $+$ 2{,}230 complex-change samples (6{,}230 total). This balanced coverage prevents the model from developing pathological biases toward any single difficulty tier. A full composition ablation comparing alternative data mixtures is provided in Table~\ref{tab:sft_composition} (Appendix~\ref{app:composition_ablation}). Balance2 achieves 45.3\% overall CM and 79.5\% FM, \textbf{surpassing all frontier LLMs on CM} ($+3.0$~pp over Qwen3.5-397B-A17B) and nearly matching the best FM (79.5\% vs.\ 80.5\% for GLM-5). The advantage is concentrated in the deployable regime: Balance2 leads by up to $+10$~pp at $|\Delta| \in \{1,2,3,4\}$, while all approaches converge near zero at $|\Delta| \geq 5$. Figure~\ref{fig:cliff_and_sft} visualizes Balance2's per-$|\Delta|$ advantage over all frontier models across the Simple and Medium tiers.

\paragraph{Downstream Validation: EnvScaler Pipeline.}
We integrate Full-Balance2 into the EnvScaler synthesis pipeline in place of its large-model ensemble. Synthesized environments are retained only if their pass accuracy meets a quality threshold of 0.85. The original pipeline yields 191 passing environments; Full-Balance2 yields 204, a 6.8\% improvement in synthesis yield, with approximately $59\times$ lower parameter count. This demonstrates that targeted specialization at the 4B scale is Pareto-superior to frontier-model pipelines on both cost and synthesis quality.

\section{Discussion and Conclusion}
\label{sec:conclusion}
 
\textbf{Why is environment simulation hard?}
Our results distinguish feedback-format errors from state-transition errors. \emph{Format mismatch} generates structurally different feedback strings that fail exact matching despite correct semantics, motivating semantic metrics as future complements to FM. \emph{State tracking failure} affects all models: predicting which nested fields change and to what values requires faithful execution of the tool's logic, a capability closer to program execution than to language modeling. Our SFT approach partially addresses this, but cases with $|\Delta|\geq 5$ remain challenging.
 
\textbf{Cost-quality trade-off.}
Full-Balance2 surpasses all frontier LLMs on CM (45.3\% vs.\ 42.3\%) and nearly matches the best FM (79.5\% vs.\ 80.5\%), while running at dramatically lower cost, both on the benchmark and in the downstream synthesis pipeline. This demonstrates that EnvSimBench enables principled cost-quality comparison and targeted model improvement.
 
\textbf{Limitations.}
\textbf{(1) Benchmark sample size and statistical reliability.}
Fine-grained per-$|\Delta|$ analysis in the Difficult group rests on strata as small as $n=4$--$6$, where a single sample shifts CM by 17--25 pp. The Failure group ($n=20$) shows the same fragility. Deterministic ground-truth labels do not eliminate selection-induced sampling variance, and bootstrap error bounds should accompany all quantitative findings in future work.
\textbf{(2) Circular downstream evaluation.}
The synthesis quality filter overlaps in distribution with SFT training data, introducing circular validation. Scarcity of high-$|\Delta|$ training samples further compounds this; targeted data augmentation for $|\Delta| \geq 5$ remains the primary open challenge.

\textbf{Conclusion.} We introduced EnvSimBench, a benchmark that reframes LLM environment simulation as a fully observable state prediction task, enabling objective, LLM-free evaluation across 400 samples and 167 diverse tool-interactive environments. Our central finding, the state-change cliff, is a pronounced but non-monotonic decline in configuration accuracy on state-changing operations. The constraint-driven formulation makes the recorded state and transition logic explicit, and targeted fine-tuning produces a 4B model that surpasses the evaluated non-thinking frontier baselines on Config Match. We release EnvSimBench as a public diagnostic framework and hope it grounds future research on reliable environment simulation, a prerequisite for trustworthy scalable agent training.

 
\subsection*{AI Use Statement}
Generative AI assisted with prose revision, result-table formatting, visualization code, and consistency checks against the supplied manuscript, implementation, figures, and experiment records.

\subsection*{Ethics Statement}
This research adheres to the ICLR Code of Ethics. This work introduces EnvSimBench, a benchmark for evaluating environment simulation capabilities of large language models, and proposes a constraint-driven simulation paradigm to improve simulation reliability. Our research does not involve human subjects, animal testing, or the collection or use of personally identifiable information. All experiments are performed using publicly accessible LLM inference APIs and synthetically constructed tool-interactive environments generated by the EnvScaler pipeline. No sensitive or private data is included in the benchmark or training data.

\subsection*{Reproducibility Statement}
We have made substantial efforts to ensure the reproducibility of all results presented in this paper. The complete benchmark construction methodology, evaluation protocols, and metric definitions are described in detail throughout the main text and in Appendix~B (Benchmark Construction Pipeline) and Appendix~G (Experimental Setup Details). This includes the three-axis difficulty stratification strategy, the diversity rule for sample selection, and the formal specifications of Feedback Match and Config Match. All model identifiers, API endpoints, training hyperparameters, and inference configurations are explicitly documented in Appendix~G and Appendix~H, allowing exact replication of both frontier model evaluations and SFT training experiments. The full EnvSimBench benchmark code, dataset, and evaluation scripts are publicly available at \url{https://github.com/cookieApril/EnvSimBench}. The release includes all 400 benchmark samples with Python-executed ground truth labels, the constraint-driven simulation pipeline, and all scripts needed to reproduce the figures and quantitative results. Supplementary results, ablation studies, and expanded analyses provided in the appendices can all be reproduced using the released code and data.

\bibliographystyle{iclr2027_conference}
\bibliography{iclr2027_conference}

\clearpage
\appendix

\newcommand{\cmark}{\ensuremath{\checkmark}}
\newcommand{\xmark}{\ensuremath{\times}}

\section{Detailed Per-Delta Experimental Results}
\label{app:detailed_results}

\subsection{\texorpdfstring{Simple Group ($\Delta \in \{1,2\}$)}{Simple Group}}

\begin{table}[htbp]
\centering
\small
\setlength{\tabcolsep}{4pt}
\caption{Results on Simple Group.
  \textbf{Bold}: column-wise maximum;
  \textcolor{red}{red}: column-wise minimum.}
\begin{tabular}{l cc cc cc}
\toprule
& \multicolumn{2}{c}{$\Delta=1$} & \multicolumn{2}{c}{$\Delta=2$} & \multicolumn{2}{c}{Average} \\
\cmidrule(lr){2-3}\cmidrule(lr){4-5}\cmidrule(lr){6-7}
Model & FM & CM & FM & CM & FM & CM \\
\midrule
DeepSeek-V3.2
  & 80\% & \textcolor{red}{36\%}
  & 80\% & \textcolor{red}{8\%}
  & 80\% & \textcolor{red}{22\%} \\
Qwen3.5-397B-A17B
  & 84\% & 64\%
  & 84\% & 28\%
  & 84\% & 46\% \\
GPT-5.4
  & 84\% & 60\%
  & 76\% & 20\%
  & 80\% & 40\% \\
Gemini-3.1-Pro-Preview
  & \textbf{100\%} & 68\%
  & \textbf{92\%}  & 28\%
  & \textbf{96\%}  & 48\% \\
Claude-Sonnet-4.6
  & \textcolor{red}{8\%}  & 56\%
  & \textcolor{red}{12\%} & 24\%
  & \textcolor{red}{10\%} & 40\% \\
MiniMax-M2.7
  & 40\% & 72\%
  & 24\% & 28\%
  & 32\% & 50\% \\
GLM-5
  & 96\% & 64\%
  & \textbf{92\%} & 24\%
  & 94\% & 44\% \\
\midrule
Full-Balance2
  & 96\%          & \textbf{80\%}
  & 88\%          & \textbf{36\%}
  & 92\%          & \textbf{58\%} \\
\bottomrule
\end{tabular}
\end{table}

\subsection{\texorpdfstring{Medium Group ($\Delta \in \{3,4,5,6\}$)}{Medium Group}}

\begin{table}[htbp]
\centering
\small
\setlength{\tabcolsep}{3.5pt}
\caption{Results on Medium Group.
  \textbf{Bold}: column-wise maximum;
  \textcolor{red}{red}: column-wise minimum.}
\begin{tabular}{l cc cc cc cc cc}
\toprule
& \multicolumn{2}{c}{$\Delta=3$} & \multicolumn{2}{c}{$\Delta=4$} &
  \multicolumn{2}{c}{$\Delta=5$} & \multicolumn{2}{c}{$\Delta=6$} &
  \multicolumn{2}{c}{Avg} \\
\cmidrule(lr){2-3}\cmidrule(lr){4-5}\cmidrule(lr){6-7}\cmidrule(lr){8-9}\cmidrule(lr){10-11}
Model & FM & CM & FM & CM & FM & CM & FM & CM & FM & CM \\
\midrule
DeepSeek-V3.2
  & 86\% & \textcolor{red}{24\%}
  & 88\% & \textcolor{red}{6\%}
  & 74\% & \textbf{4\%}
  & 58\% & \textcolor{red}{0\%}
  & 76.5\% & \textcolor{red}{8.5\%} \\
Qwen3.5-397B-A17B
  & 78\% & 30\%
  & 62\% & 30\%
  & 60\% & 2\%
  & 60\% & 6\%
  & 65\%   & 17\% \\
GPT-5.4
  & 82\% & 36\%
  & 70\% & 24\%
  & 74\% & 2\%
  & \textbf{74\%} & \textbf{8\%}
  & 75\%   & 17.5\% \\
Gemini-3.1-Pro-Preview
  & 90\% & 36\%
  & 74\% & 24\%
  & 80\% & \textbf{4\%}
  & 28\% & 6\%
  & 68\%   & 17.5\% \\
Claude-Sonnet-4.6
  & \textcolor{red}{0\%}  & 30\%
  & \textcolor{red}{0\%}  & 16\%
  & \textcolor{red}{22\%} & \textcolor{red}{0\%}
  & 28\% & 2\%
  & \textcolor{red}{12.5\%} & 12\% \\
MiniMax-M2.7
  & 24\% & 28\%
  & 28\% & 26\%
  & 26\% & 2\%
  & \textcolor{red}{10\%} & \textbf{8\%}
  & 22\%   & 16\% \\
GLM-5
  & 90\% & 28\%
  & 86\% & 22\%
  & 90\% & 2\%
  & 60\% & 4\%
  & 81.5\% & 14\% \\
\midrule
Full-Balance2
  & \textbf{92\%} & \textbf{44\%}
  & \textbf{98\%} & \textbf{40\%}
  & \textbf{94\%} & 2\%
  & \textbf{74\%} & \textbf{8\%}
  & \textbf{89.5\%} & \textbf{23.5\%} \\
\bottomrule
\end{tabular}
\end{table}

\subsection{\texorpdfstring{Difficult Group ($\Delta \in \{7,\ldots,12\}$)}{Difficult Group}}
 
\begin{table}[htbp]
\centering
\small
\setlength{\tabcolsep}{2.8pt}
\caption{%
  Results on Difficult Group, per $\Delta$ value and averaged.
  \textbf{Bold}: column-wise maximum; \textcolor{red}{red}: column-wise minimum.
  The CM rebound at $\Delta=7$ reflects bulk-uniform operations
  (see Appendix~\ref{app:difficult_pattern})
}
\label{tab:difficult_perdelta}
\begin{tabular}{l cc cc cc cc cc cc cc}
\toprule
& \multicolumn{2}{c}{$\Delta=7$}
& \multicolumn{2}{c}{$\Delta=8$}
& \multicolumn{2}{c}{$\Delta=9$}
& \multicolumn{2}{c}{$\Delta=10$}
& \multicolumn{2}{c}{$\Delta=11$}
& \multicolumn{2}{c}{$\Delta=12$}
& \multicolumn{2}{c}{\textbf{Avg}} \\
\cmidrule(lr){2-3}\cmidrule(lr){4-5}\cmidrule(lr){6-7}
\cmidrule(lr){8-9}\cmidrule(lr){10-11}\cmidrule(lr){12-13}\cmidrule(lr){14-15}
\textbf{Model}
  & FM & CM & FM & CM & FM & CM
  & FM & CM & FM & CM & FM & CM
  & FM & CM \\
\midrule
DeepSeek-V3.2
  & 19\% & \textcolor{red}{6\%} & 25\% & \textcolor{red}{0\%} & 20\% & \textcolor{red}{0\%} & 55\% & \textcolor{red}{9\%} & \textcolor{red}{0\%} & \textcolor{red}{0\%} & 25\% & \textcolor{red}{0\%} & 26\% & \textcolor{red}{4\%} \\
Qwen3.5-397B-A17B
  & 63\% & 38\% & 38\% & \textbf{25\%} & \textcolor{red}{0\%} & \textcolor{red}{0\%} & 45\% & \textbf{18\%} & 33\% & \textbf{17\%} & 25\% & \textbf{25\%} & 42\% & 24\% \\
GPT-5.4
  & \textbf{88\%} & 44\% & \textbf{100\%} & \textbf{25\%} & \textbf{60\%} & \textcolor{red}{0\%} & \textbf{73\%} & \textbf{18\%} & 50\% & \textbf{17\%} & 25\% & \textbf{25\%} & \textbf{74\%} & 26\% \\
Gemini-3.1-Pro-Preview
  & 63\% & 31\% & 88\% & 13\% & 40\% & \textcolor{red}{0\%} & 64\% & \textbf{18\%} & \textbf{83\%} & \textcolor{red}{0\%} & \textbf{75\%} & \textbf{25\%} & 68\% & 18\% \\
Claude-Sonnet-4.6
  & 19\% & 13\% & 25\% & \textbf{25\%} & \textcolor{red}{0\%} & \textcolor{red}{0\%} & \textcolor{red}{9\%} & \textbf{18\%} & \textcolor{red}{0\%} & \textbf{17\%} & \textcolor{red}{0\%} & \textbf{25\%} & 12\% & 16\% \\
MiniMax-M2.7
  & \textcolor{red}{13\%} & \textbf{50\%} & \textcolor{red}{0\%} & \textcolor{red}{0\%} & \textcolor{red}{0\%} & \textcolor{red}{0\%} & 18\% & \textbf{18\%} & \textcolor{red}{0\%} & \textcolor{red}{0\%} & \textcolor{red}{0\%} & \textbf{25\%} & \textcolor{red}{8\%} & 22\% \\
GLM-5
  & 75\% & 44\% & 63\% & \textbf{25\%} & \textbf{60\%} & \textbf{20\%} & \textbf{73\%} & \textbf{18\%} & \textbf{83\%} & \textbf{17\%} & 50\% & \textbf{25\%} & 70\% & \textbf{28\%} \\
\bottomrule
\end{tabular} 
\end{table}

\subsection{Non-Monotonicity at $|\Delta|\geq 7$: Bulk-Uniform Operations}
\label{app:difficult_pattern}

Per-$|\Delta|$ results show a partial CM recovery at $|\Delta|=7$, where Gemini~3.1~Pro, GLM-5, MiniMax-M2.7, and GPT-5.4 reach 31--50\% CM before collapsing again at $|\Delta|\geq 8$. This rebound is not evidence of improved state-tracking capability. Inspection of the 15 Difficult-group samples on which at least one model achieves CM\,=\, True reveals a consistent structural regularity. Every such sample corresponds to a \emph{bulk-uniform} operation, in which a single tool call produces $N$ structurally identical changes under the same path prefix. Two dominant patterns account for nearly all recoveries.

\textbf{Pattern 1: Consecutive date-entry additions.} A tool call such as \texttt{block\_dates(property\_id, start, end)} adds one availability record per day across a date range, yielding $|\Delta|=N$ changes that all follow an identical template: \texttt{availability\_calendar.\{property\_id\}.\{YYYY-MM-DD\}}$\leftarrow$\texttt{\{status: blocked\}}, with keys that are consecutive calendar dates. Sample~376 ($|\Delta|=10$) exemplifies this: all six models answer correctly, because identifying the pattern ``add blocked entries for 2025-05-01 through 2025-05-10'' reduces the task to enumerating a date sequence. No independent value derivation is required per entry. Samples~397 ($|\Delta|=12$) and~352 ($|\Delta|=7$) follow the same structure and achieve 5/6 and 5/6 correct respectively.

\textbf{Pattern 2: Multi-field update on a single object.} A call that overwrites all fields of one record (e.g., \texttt{update\_reservation}) produces $|\Delta|$ equal to the number of updatable fields. Still, every change concerns the same object (e.g., \texttt{reservations.RES2.*}) and its values are read directly from the call arguments. Sample~377 ($|\Delta|=10$) achieves 5/6 correct for exactly this reason: no cross-object inference or runtime-value derivation is required.

The contrast with Medium-group failures is instructive. A $|\Delta|=5$ sample such as \texttt{verify\_otp} requires five heterogeneous updates across three sub-objects, including a clock-generated timestamp whose value is unknowable without executing the transition logic. Each field demands an independent derivation. Bulk-uniform samples collapse this complexity into a single template instantiated $N$ times, structurally easier despite the larger $|\Delta|$.

This distinction has a direct implication for benchmark design. The Difficult tier currently conflates two qualitatively different difficulty sources: \emph{heterogeneous multi-field} updates, which cause a permanent CM collapse, and \emph{homogeneous bulk} updates, which remain tractable at high $|\Delta|$. Future iterations of EnvSimBench should stratify these two sub-types separately to provide more precise diagnostic resolution at the upper end of the complexity axis.
\section{Benchmark Construction Pipeline (Detailed)}
\label{app:construction}

\begin{figure*}[htbp]
    \centering
    \includegraphics[width=0.92\linewidth]{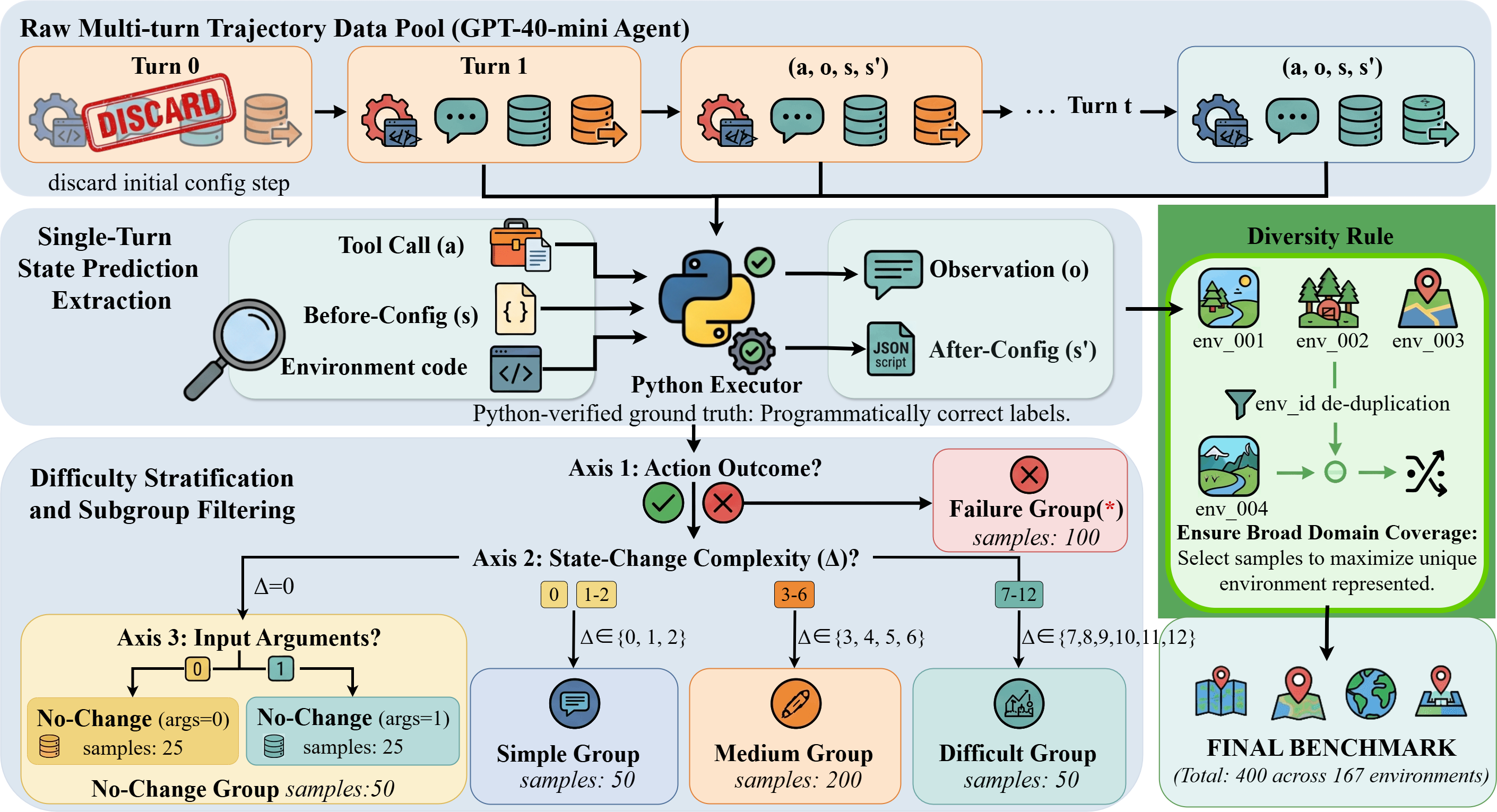}
    \caption{%
        \textbf{Benchmark construction pipeline.}
        \emph{Left}: Multi-turn trajectories from 191 EnvScaler environments are collected by a GPT-4o-mini agent and preprocessed into single-turn $(a,o,s,s')$ samples; Step~0 (no before-config) is discarded.
        \emph{Center}: Three-axis stratification partitions the pool by action outcome, state-change complexity, and input argument cardinality.
        \emph{Right}: A diversity rule maximizes distinct \texttt{env\_id}s per subgroup, yielding 400 samples across 167 environments with Python-executed ground truth.
    }
    \label{fig:benchmark_construction}
\end{figure*}

Figure~\ref{fig:benchmark_construction} illustrates the full sample-selection pipeline with three-axis stratification and the diversity rule—the pipeline proceeds in three stages.

\textbf{Stage 1: trajectory collection.} GPT-4o-mini is deployed as an agent across all 191 EnvScaler environments, executing up to 30 steps per environment. At each step $t\geq1$, the Python executor records the complete quadruple $(a, o, s, s')$: the tool call, the resulting observation, the before-config, and the after-config, along with the derived change set $\Delta(s,s')$. The initialization-only record is discarded because it has no corresponding tool-call transition. This yields a large pool of candidate single-turn state prediction samples.

\textbf{Stage 2: three-axis stratification and filtering.} Candidate samples are classified simultaneously along the three axes described in Section~\ref{sec:bench}. Axis~1 partitions by action outcome (\texttt{success=True/False}); Axis~2 partitions the success group by $|\Delta|$ into No-Change, Simple, Medium, and Difficult; Axis~3 further splits Failure and No-Change subgroups by input argument cardinality (0 vs.\ 1). Each resulting subgroup is then subsampled to its target count (Table~\ref{tab:sample_distribution}).

\textbf{Stage 3: diversity rule.} Within each subgroup, samples are selected greedily to maximize the number of distinct \texttt{env\_id}s covered. If the target count exceeds the number of available environments in the subgroup, multiple samples per environment are permitted, prioritizing those with different $|\Delta|$ values. This rule ensures that the 400 benchmark samples span 167 distinct environments rather than being concentrated in a few high-trajectory environments, providing broad domain coverage for evaluation.

\section{Performance Visualizations}
\label{app:figures}

\begin{figure*}[htbp]
    \centering
    \includegraphics[width=\textwidth]{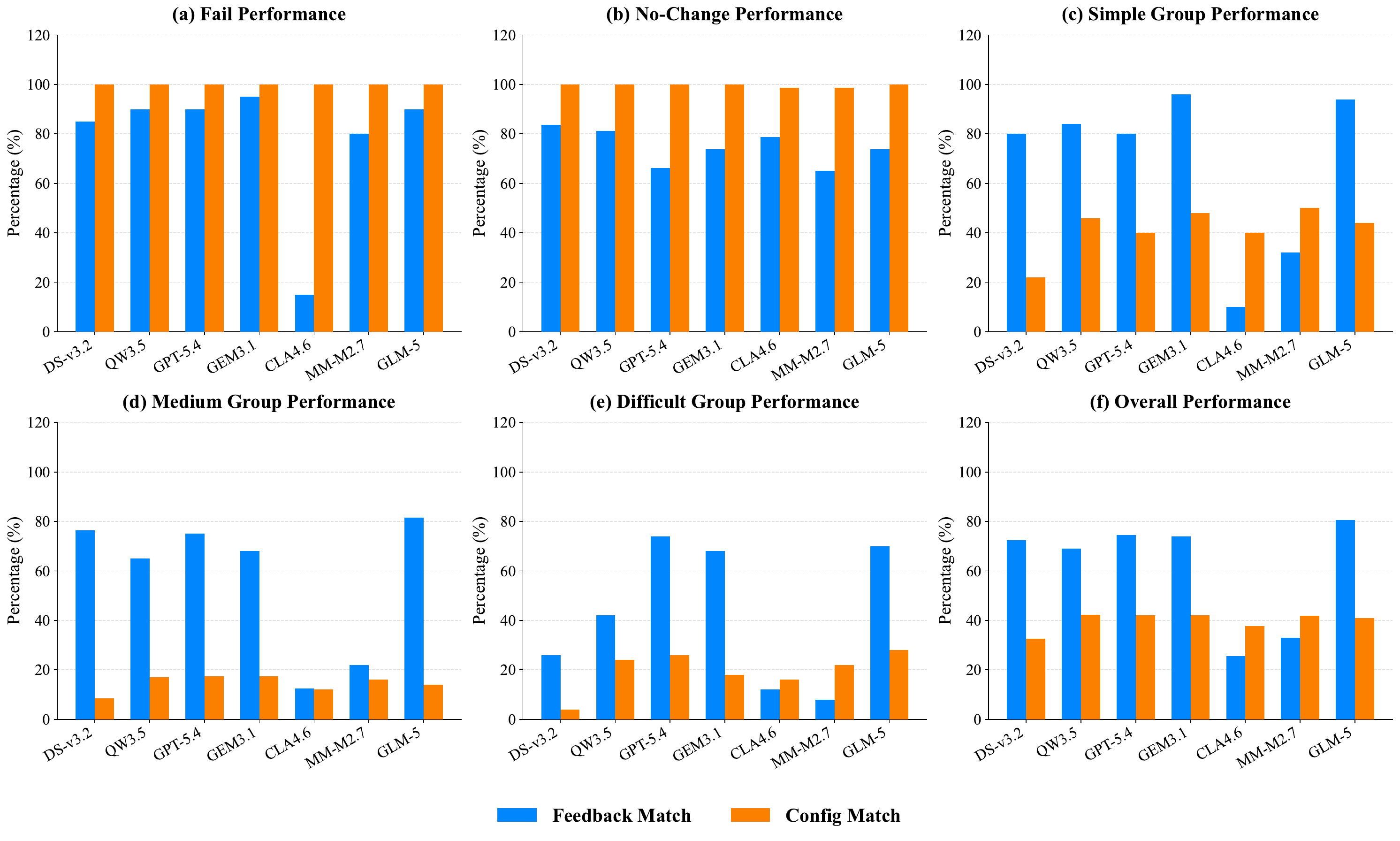}
    \caption{%
    \textbf{Frontier LLM performance across difficulty subgroups.}
    Blue bars show FM and orange bars show CM for seven frontier
    models under non-thinking inference on the original benchmark.
    Panels~(a)--(b): CM is near-perfect on Failure and No-Change samples,
    while FM varies across models.
    Panels~(c)--(e): CM ranges from 22--50\% on Simple samples,
    8.5--17.5\% on Medium samples, and 4--28\% on Difficult samples.
    Panel~(f): GLM-5 achieves the highest overall FM (80.5\%),
    while Qwen3.5-397B-A17B achieves the highest overall CM (42.3\%).
}
    \label{fig:performance_bars}
\end{figure*}

\subsection{Bar Chart Analysis: Per-Group FM and CM (Figure~\ref{fig:performance_bars})}

Figure~\ref{fig:performance_bars} renders FM (blue bars) and CM (orange bars) side by side for all seven frontier models across five subgroups and the overall benchmark.

\textbf{Panels (a) Failure and (b) No-Change: high CM with variable FM.} All seven models achieve 100\% CM on Failure samples, while CM on the pooled No-Change group ranges from 98.75\% to 100\%. Feedback accuracy varies more widely. Claude-Sonnet-4.6 achieves 15\% FM on Failure samples despite 100\% CM, illustrating that feedback errors can occur even when the predicted change set is correct. On the pooled No-Change group, GPT-5.4 achieves 66.25\% FM and MiniMax-M2.7 achieves 65\%, the lowest score in this group. The examples in Appendix~\ref{app:failure_analysis} distinguish output-format mismatches from feedback-content errors.

\textbf{Panel (c) Simple ($|\Delta|\in\{1,2\}$): different rankings under FM and CM.}
FM ranges from 10\% to 96\%, while CM ranges from 22\% to 50\%.
Gemini-3.1-Pro-Preview achieves the highest FM (96\%), whereas
MiniMax-M2.7 achieves the highest CM (50\%), followed by Gemini
at 48\%. FM exceeds CM for five of the seven models.
The exceptions are Claude-Sonnet-4.6 (10\% FM versus 40\% CM)
and MiniMax-M2.7 (32\% versus 50\%).
DeepSeek-V3.2 exhibits a large gap in the opposite direction,
with 80\% FM and 22\% CM.
These differences show that feedback accuracy and configuration
accuracy induce different model rankings even on Simple samples.

\textbf{Panels (d) Medium ($|\Delta|\in\{3,\ldots,6\}$) and
(e) Difficult ($|\Delta|\in\{7,\ldots,12\}$): low CM with variation across groups.}
On Medium samples, CM ranges from 8.5\% to 17.5\%, while FM
ranges from 12.5\% to 81.5\%.
DeepSeek-V3.2 and GLM-5 exhibit the largest FM--CM gaps in this
group, at 68.0 and 67.5 percentage points, respectively.
On Difficult samples, CM ranges from 4\% to 28\%.
GPT-5.4 achieves 74\% FM and 26\% CM, while GLM-5 achieves
70\% FM and 28\% CM.
CM remains below the Simple-group score for every model,
but several models score higher on Difficult than on Medium.
The group-level results therefore show low configuration accuracy
on complex transitions with a non-monotonic relationship to
state-change count.

\textbf{Panel (f) Overall: model ranking summary.}
The aggregate view makes the cross-model contrast most accessible. GLM-5 has the tallest FM bar (80.5\%) but a moderate CM bar (41.0\%). Qwen3.5-397B-A17B has the tallest CM bar (42.3\%). The two format-divergent models, Claude-Sonnet-4.6 and MiniMax-M2.7, have the shortest FM bars despite having CM bars comparable to the other models. Low FM does not imply low CM.

\begin{figure*}[htbp]
    \centering
    \includegraphics[width=\textwidth]{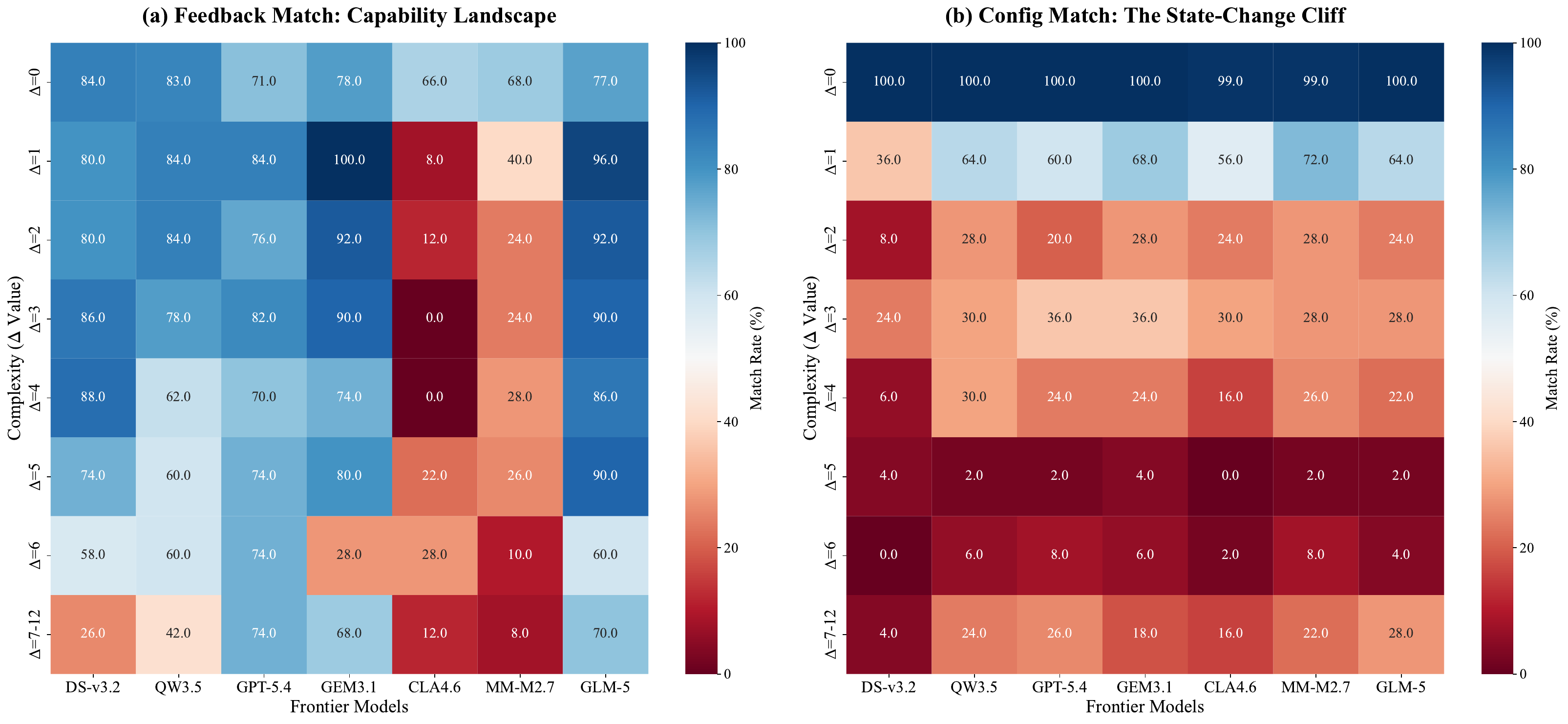}
    \caption{\textbf{Feedback and configuration accuracy by state-change count.}
The heatmaps report FM (left) and CM (right) for seven frontier models under non-thinking inference. The $|\Delta|=0$ row pools Failure and No-Change samples, and the final row pools $|\Delta|=7$--12. CM is 99--100\% at $|\Delta|=0$ and 0--4\% at $|\Delta|=5$. Higher feedback accuracy can coexist with low configuration accuracy at the same change count.}
    \label{fig:heatmaps}
\end{figure*}

\subsection{Heatmap Analysis: FM and CM by State-Change Count
(Figure~\ref{fig:heatmaps})}

Figure~\ref{fig:heatmaps} reports FM on the left and CM on the
right for seven frontier models under non-thinking inference.
Columns correspond to models, and rows correspond to state-change
counts $|\Delta|$. The $|\Delta|=0$ row pools Failure and No-Change
samples, while the final row pools $|\Delta|\in\{7,\ldots,12\}$.
Blue indicates higher match rates and red indicates lower match rates.

\textbf{Left panel (FM): variation across models and state-change counts.}
Claude-Sonnet-4.6 and MiniMax-M2.7 achieve 66\% and 68\% FM,
respectively, on the pooled $|\Delta|=0$ group.
Their FM is lower on every state-changing row: Claude ranges
from 0\% to 28\%, and MiniMax ranges from 8\% to 40\% across
the displayed rows.
The other five models also vary across change counts.
For example, Gemini-3.1-Pro-Preview achieves 100\% FM at
$|\Delta|=1$, 28\% at $|\Delta|=6$, and 68\% on the pooled
$|\Delta|=7$--12 group.
Thus, feedback accuracy varies across both models and sample strata.
The format and content errors underlying these differences are
examined through individual examples in
Appendix~\ref{app:failure_analysis}.

\textbf{Right panel (CM): low accuracy on multi-field transitions.}
CM is 99--100\% on the pooled $|\Delta|=0$ group and ranges
from 36\% to 72\% at $|\Delta|=1$.
It falls to 8--28\% at $|\Delta|=2$, but rises to 24--36\%
at $|\Delta|=3$.
The lowest ranges occur at $|\Delta|=5$ and $|\Delta|=6$,
where CM is 0--4\% and 0--8\%, respectively.
On the pooled $|\Delta|=7$--12 group, CM ranges from 4\%
to 28\%.
These results show a pronounced loss of configuration accuracy
relative to state-preserving transitions, with a non-monotonic
pattern across state-change counts.
MiniMax-M2.7 leads CM on the Simple group (50\%), while GPT-5.4
and Gemini-3.1-Pro-Preview tie for the highest Medium-group CM (17.5\%).

\textbf{Comparing the two panels jointly.}
At $|\Delta|=5$, GLM-5 achieves 90\% FM and 2\% CM,
while DeepSeek-V3.2 achieves 74\% FM and 4\% CM.
These gaps show that high feedback accuracy can coexist with
low configuration accuracy within the same sample stratum.
Feedback accuracy alone therefore does not establish reliable
state prediction.
Appendix~\ref{app:failure_analysis} reports the instance-level
analysis of correct feedback paired with incorrect state changes.

\section{Extended Failure Mode Analysis}
\label{app:failure_analysis}

This appendix provides expanded qualitative and quantitative analyses of the three
state-tracking failure sub-patterns identified in Finding~3, together with cross-model comparisons on shared benchmark samples.

\begin{table}[htbp]
\centering
\small
\setlength{\tabcolsep}{4pt}
\caption{Format divergence examples. CM\,=\,100\% in all listed cases; FM\,=\,0\%.}
\label{tab:format_examples}
\begin{tabular}{p{1.1cm} p{4.5cm} p{4.5cm}}
\toprule
\textbf{Model} & \textbf{Predicted feedback} & \textbf{Ground-truth feedback} \\
\midrule
Claude
  & \texttt{Security not found}
  & \texttt{\{'success': False, 'error': 'Security not found'\}} \\
\addlinespace
Claude
  & \texttt{No observations found for station 'ST001' on 2024-04-19.}
  & \texttt{\{'success': False, 'error': 'No observations found\ldots'\}} \\
\addlinespace
Claude
  & \texttt{Exactly one of 'delta\_memory' or 'multiplier' must be provided.}
  & \texttt{\{'success': False, 'error': 'Exactly one of\ldots'\}} \\
\addlinespace
MiniMax
  & \texttt{User reputation updated successfully.}
  & \texttt{\{'success': True, 'message': 'User reputation\ldots'\}} \\
\addlinespace
MiniMax
  & \texttt{Movie 'tt0111161' updated successfully.}
  & \texttt{\{'success': True, 'message': "Movie 'tt0111161'\ldots"\}} \\
\bottomrule
\end{tabular}
\end{table}

\subsection{Format Divergence: Full Example Set}
\label{app:format_divergence}

As described in Finding~3, Claude-Sonnet-4.6 and MiniMax-M2.7 systematically unwrap the structured response schema and return only the inner message string. Table~\ref{tab:format_examples} enumerates representative cases. In every instance, CM is 100\% (state prediction is correct) while FM is 0\% (string mismatch). These examples demonstrate format sensitivity in FM; they do not establish a uniform error pattern across all difficulty levels.

One edge case deserves attention: a second Claude failure-group example predicts \texttt{"Workout scheduled successfully"} where the ground truth is a failure response. Here, beyond the format mismatch, the model also misclassifies the error condition. The rare case where FM\,=\,0 reflects a genuine semantic error rather than a pure formatting convention difference.

This pattern reveals a counterintuitive deployment tradeoff: Claude-Sonnet-4.6 and MiniMax-M2.7, despite their low absolute FM, may be \emph{safer} choices for simulation pipelines that use feedback mismatches as a reliability signal, since their errors are more likely to be \emph{visible} to the monitoring layer. In contrast, models with near-perfect FM but poor CM produce silent corruptions that accumulate undetected.

\subsection{State-Change Cliff: Cross-Model Analysis on Shared Samples}
\label{app:cliff_examples}

\paragraph{Example 1: \texttt{verify\_otp} ($|\Delta|=5$).}
This call must update five fields across three sub-objects simultaneously.
Table~\ref{tab:verifyotp_example} shows all seven models' field-level predictions on the same sample. Ground truth requires all five fields to be updated.

\begin{table}[htbp]
\centering
\small
\setlength{\tabcolsep}{5pt}
\caption{%
  Cross-model field-level predictions on \texttt{verify\_otp} ($|\Delta|=5$).
  \cmark\,=\,correctly predicted; \xmark\,=\,missed or wrong value;
  $\dagger$\,=\,path correctly identified but execution-time value wrong
  (stale-approximation error, sub-type~(b) of Appendix~\ref{app:timestamp_analysis}).
  Ground truth: all five fields must be updated.
}
\label{tab:verifyotp_example}
\begin{tabular}{l ccccc cc}
\toprule
\textbf{Model} & \textbf{is\_ver.} & \textbf{timestamp} & \textbf{attempt}
& \textbf{is\_used} & \textbf{sess.} & \textbf{FM} & \textbf{CM} \\
\midrule
DeepSeek-V3.2        & \xmark & \xmark          & \cmark & \cmark & \cmark & \cmark & \xmark \\
Qwen3.5-397B-A17B    & \cmark & $\dagger$\xmark  & \cmark & \cmark & \cmark & \cmark & \xmark \\
Gemini-3.1-Pro-Preview       & \xmark & \xmark          & \cmark & \cmark & \cmark & \cmark & \xmark \\
GLM-5                & \xmark & \xmark          & \cmark & \cmark & \cmark & \cmark & \xmark \\
MiniMax-M2.7         & \cmark & \xmark          & \cmark & \cmark & \xmark & \cmark & \xmark \\
Claude-Sonnet-4.6    & \cmark & \xmark          & \xmark & \cmark & \cmark & \cmark & \xmark \\
GPT-5.4              & \xmark & \xmark          & \xmark & \xmark & \xmark & \xmark & \xmark \\
\bottomrule
\end{tabular}
\end{table}

Three patterns are salient. First, the runtime-generated timestamp field is missed by \emph{every} model: set to the current epoch at call time, it cannot be inferred statically. Second, DeepSeek-V3.2, Gemini-3.1-Pro-Preview, and GLM-5 independently converge on the same three-field prediction, suggesting a shared attentional bias toward the most linguistically salient fields. Third, GPT-5.4 collapses to zero predicted changes despite generating a feedback string, the full cliff failure mode. A fourth pattern distinguishes Qwen3.5-397B-A17B from the other six: it is the only model to identify all five field paths, yet still fails CM because its predicted \texttt{last\_verification\_timestamp} (\texttt{1774167900.0}) differs from the executor's runtime value (\texttt{1774167865.92}). This is a stale-approximation error rather than a path-identification failure, placing it in sub-type~(b) of Appendix~\ref{app:timestamp_analysis}.

\paragraph{Example 2: \texttt{return\_equipment} ($|\Delta|=5$).}
The ground-truth transition requires four primary field updates plus one newly created auxiliary log record. Every model fails:

\begin{itemize}
\item \textbf{DeepSeek-V3.2}: correctly updates three primary fields; omits one primary field and the auxiliary log record. FM\,=\,True, CM\,=\,False.
\item \textbf{Claude-Sonnet-4.6}: assigns a domain-plausible but code-absent value to the primary status field; omits the auxiliary record. FM\,=\,False, CM\,=\,False.
\item \textbf{Gemini-3.1-Pro-Preview / GLM-5 / MiniMax-M2.7}: same incorrect status value as Claude; omit one primary field and the auxiliary record. FM\,=\,False, CM\,=\,False.
\item \textbf{GPT-5.4}: correctly derives all four primary field values, the only model to do so, but fabricates a runtime-generated key for the auxiliary record that mismatches the executor's output. FM\,=\,True, CM\,=\,False.
\end{itemize}

Across all $|\Delta| \geq 3$ CM failures, models omit newly \emph{added} components in 68--79\% of cases where such additions exist, substantially exceeding the omission rate for \emph{modified} components. Four models independently hallucinate the same domain-plausible but code-absent status value, illustrating how parametric world knowledge overrides code-grounded reasoning when the required value is not directly readable from $\texttt{code}(a_t)$.

\subsection{Runtime-Dependent Values: Failure Taxonomy}
\label{app:timestamp_analysis}

Timestamp-bearing and runtime-generated fields appear in the ground truth of 23--28\% of all $|\Delta| \geq 3$ CM failures across models (Table~\ref{tab:timestamp_stats}). A model performing static reasoning over $\texttt{code}(a_t)$ cannot infer execution-time values without actually running the transition logic; it must either omit the field entirely or substitute a plausible but incorrect prior value.
Three sub-patterns are observed.

\begin{table}[htbp]
\centering
\small
\caption{Presence of runtime-dependent fields in $|\Delta|\geq3$ CM-failure samples.}
\label{tab:timestamp_stats}
\begin{tabular}{lcc}
\toprule
\textbf{Model} & \textbf{CM failures ($|\Delta|\geq3$)} & \textbf{With runtime field} \\
\midrule
DeepSeek-V3.2     & 231 & 24\% \\
Gemini-3.1-Pro-Preview    & 206 & 23\% \\
GPT-5.4           & 202 & 28\% \\
GLM-5             & 208 & 27\% \\
Claude-Sonnet-4.6 & 218 & 26\% \\
MiniMax-M2.7      & 207 & 28\% \\
\bottomrule
\end{tabular}
\end{table}

\textbf{(a) Complete omission.} The model identifies the correct field path but omits it from $\widehat{\Delta}$ entirely. This is the dominant pattern for DeepSeek-V3.2 and GLM-5.

\textbf{(b) Stale-value substitution.} The model includes the field with a plausible but stale value interpolated from the before-state $s_t$. In a representative case, a model predicts a stale proxy $\hat{v}_\tau$ for a clock-dependent field while the executor writes the true runtime value $v^*_\tau \neq \hat{v}_\tau$; the surface response $\hat{o}_t = o_t$ nonetheless matches (FM\,=\,1, CM\,=\,0).

\textbf{(c) Transaction-ID key mismatch.} For auto-generated record keys encoding a runtime epoch, a model must both create the record \emph{and} predict the correct key. A correct record body with a wrong key still fails CM, a compound challenge of hidden side effect plus runtime dependency, as illustrated by GPT-5.4 in Example~2 above.

\subsection{Plausible Feedback Masking Wrong State: Quantitative Breakdown}
\label{app:fm_true_cm_false}

Table~\ref{tab:fm_t_cm_f} reports the fraction of CM-failure samples that also pass FM at each $|\Delta|$ level. These are the most dangerous cases for simulation-as-training: the agent's reward signal is intact while the environment state is silently wrong.

\begin{table}[htbp]
\centering
\small
\setlength{\tabcolsep}{5pt}
\caption{%
  Percentage of CM-failure samples that also pass FM, by $|\Delta|$ level. High values indicate silent state corruption that is invisible to the agent.
  \textbf{Bold}: column-wise maximum;
  \textcolor{red}{red}: column-wise minimum.
}
\label{tab:fm_t_cm_f}
\begin{tabular}{lcccccc}
\toprule
\textbf{Model} & $\Delta=1$ & $\Delta=2$ & $\Delta=3$ & $\Delta=4$ & $\Delta=5$ & $\Delta=6$ \\
\midrule
DeepSeek-V3.2
  & 81\%          & 78\%
  & 82\%          & \textbf{87\%}
  & 75\%          & 58\% \\
GLM-5
  & 89\%          & \textbf{89\%}
  & \textbf{86\%} & 85\%
  & \textbf{90\%} & 58\% \\
GPT-5.4
  & 60\%          & 70\%
  & 72\%          & 63\%
  & 73\%          & \textbf{72\%} \\
Gemini-3.1-Pro-Preview
  & \textbf{100\%} & \textbf{89\%}
  & 84\%           & 66\%
  & 79\%           & 23\% \\
Claude-Sonnet-4.6
  & 18\%                  & \textcolor{red}{11\%}
  & \textcolor{red}{0\%}  & \textcolor{red}{0\%}
  & \textcolor{red}{22\%} & 29\% \\
MiniMax-M2.7
  & \textcolor{red}{14\%} & 28\%
  & 19\%                  & 35\%
  & 27\%                  & \textcolor{red}{9\%} \\
\bottomrule
\end{tabular}
\end{table}

For DeepSeek-V3.2, GLM-5, GPT-5.4, and Gemini-3.1-Pro-Preview, 60--100\% of CM failures at every $|\Delta|$ level are accompanied by FM\,=\, True: the agent sees no divergence signal. For Claude-Sonnet-4.6 and MiniMax-M2.7, the pattern inverts: their format divergence means errors are \emph{more likely} also to fail FM, making divergence visible. This reveals a counterintuitive deployment tradeoff: models with format divergence, despite low absolute FM, may be safer for simulation pipelines that use feedback mismatches as a reliability signal.

In one representative case at $|\Delta|=3$, two models independently produce the correct surface response $o_t$ but collapse a nested sub-object to a coarse scalar representation (losing field-level granularity) and substitute a stale value for the clock-dependent timestamp field: $\hat{o}_t = o_t$ but $\hat{s}'_t \neq s'_t$. A third model generates correct feedback and correctly predicts two primary fields but omits the timestamp field entirely: both pass FM and fail CM. In a training pipeline, such silent divergences accumulate and systematically bias agent behavior.

\section{Hallucination Case Studies}
\label{app:hallucination}

This appendix presents executor-verified examples of fabricated rejection conditions, unsupported entity creation, and failure to use the explicitly provided before-state. These examples illustrate errors that remain possible under single-turn, explicit-state evaluation.

\subsection{Failure Mode 1 Hallucination: Fabricating Non-Existent Constraint Violations}
\label{app:halluc_fabricated_constraint}

In this failure mode, the model invents a plausible business-rule violation to reject
a call that the executor accepts. No such constraint exists in the tool code; the model
substitutes reasoning grounded in real-world domain conventions for the actual
implementation logic.

\paragraph{Sample: \texttt{add\_observations} ($|\Delta|=3$, all models fail).}
The call submits three wildlife observations with IDs \texttt{obs\_9}, \texttt{obs\_10},
\texttt{obs\_11} for distinct observers (\texttt{OBSR02}, \texttt{OBSR05},
\texttt{OBSR04}) at timestamps \texttt{1685000000}, \texttt{1686000000},
\texttt{1687000000}, each approximately 11.6 days apart. The executor accepts all
three and creates the corresponding records. The six model outputs detailed below all fail CM, but they differ in whether they reject the call or create records with incorrect values.

\begin{table}[htbp]
\centering
\small
\setlength{\tabcolsep}{4pt}
\caption{%
  Hallucination: all six models fabricate different constraint violations to reject
  three valid observations. Ground truth: 3 records created, \texttt{success=True}.
}
\label{tab:halluc_constraint}
\begin{tabular}{p{2.2cm} p{9.0cm}}
\toprule
\textbf{Model} & \textbf{Fabricated rejection reason} \\
\midrule
DeepSeek-V3.2
  & \texttt{"0 observations added, 3 failed"}; each observation individually
    flagged as \texttt{"Duplicate observation within 1 minute"}, despite timestamps
    being $\sim$11.6 days apart. \\
\addlinespace
Gemini-3.1-Pro-Preview
  & \texttt{"0 observations added, 3 failed"}; observations flagged as
    \texttt{"Invalid latitude: 38.75"}, \texttt{"Invalid latitude: 69.3"},
    \texttt{"Invalid latitude: 27.18"}, all of them are valid WGS-84 coordinates. \\
\addlinespace
GPT-5.4
  & Same fabricated latitude-validity errors as Gemini-3.1-Pro-Preview;
    predicts zero state changes. \\
\addlinespace
GLM-5
  & Same fabricated latitude-validity errors;
    predicts zero state changes. \\
\addlinespace
Claude-Sonnet-4.6
  & Correctly outputs \texttt{"3 observations added successfully, 0 failed"}
    and creates 3 records, but uses \emph{wrong field values}
    (hallucinated observer IDs and timestamps), causing CM\,=\,False. \\
\addlinespace
MiniMax-M2.7
  & Same correct feedback as Claude, but with similarly wrong field values. \\
\bottomrule
\end{tabular}
\end{table}

Among the six models detailed in Table~\ref{tab:halluc_constraint}, four (DeepSeek-V3.2, Gemini-3.1-Pro-Preview, GPT-5.4, and GLM-5) reject the valid call using constraints absent from the implementation. Claude-Sonnet-4.6 and MiniMax-M2.7 instead report success and create the requested records, but predict incorrect field values. The example distinguishes fabricated rejection conditions from incorrect values in otherwise correctly identified updates.

\subsection{Failure Mode 2 Logical Inconsistency: Self-Contradictory Response}
\label{app:halluc_inconsistency}

In this case, the predicted feedback and state changes agree with each other but jointly deviate from the tool call: the model introduces an additional entity that the action did not request. The error is unsupported entity creation rather than a contradiction between the two predicted outputs.

\paragraph{Sample: \texttt{run\_bulk\_analysis} ($|\Delta|=3$, DeepSeek-V3.2).}
The call requests a trend analysis for one Japanese stock, \texttt{7203.T} (Toyota).
The executor processes only \texttt{7203.T} and updates three fields accordingly.
DeepSeek-V3.2 produces a response that is internally self-contradictory in two ways
simultaneously.

\begin{table}[htbp]
\centering
\small
\setlength{\tabcolsep}{4pt}
\caption{%
  Unsupported entity creation: DeepSeek-V3.2 feedback claims ``2~stocks'' were
  analyzed, but the call only specifies 1, and the predicted state invents
  a full analysis for a second stock (\texttt{SONY}) not present in the input.
}
\label{tab:halluc_inconsistency}
\begin{tabular}{p{2.3cm} p{4.3cm} p{4.3cm}}
\toprule
& \textbf{Ground truth} & \textbf{DeepSeek-V3.2 prediction} \\
\midrule
\textbf{Feedback} &
  \texttt{"Bulk analysis completed for \textbf{1} stocks in Japan."} &
  \texttt{"Bulk analysis completed for \textbf{2} stocks in Japan."} \\
\addlinespace
\textbf{State: \texttt{7203.T}} &
  \texttt{technical\_indicators.7203.T}: updated (SMA values) \newline
  \texttt{trend\_analysis.7203.T}: \{confidence: 0.8\} &
  Same fields updated correctly (\emph{this part is right}) \\
\addlinespace
\textbf{State: \texttt{SONY}} &
  \emph{Not in request; no changes} &
  \texttt{technical\_indicators.SONY}: full SMA table \newline
  \texttt{trend\_analysis.SONY}: \{confidence: 0.8, trend: up\} \newline
  \emph{(entirely fabricated)} \\
\bottomrule
\end{tabular}
\end{table}

The inconsistency operates at two levels. First, the feedback count (``2 stocks'') contradicts the input (1 stock was requested). Second, the predicted state changes include a complete analysis of \texttt{SONY} (with internally coherent SMA values, trend direction, and confidence score) that has no basis in the call arguments. The model correctly processes \texttt{7203.T} (the core prediction is right), then spontaneously hallucinates an entire second entity. An agent receiving this response would record a ghost analysis for SONY in the environment state, causing all subsequent reasoning about the SONY data to be grounded in fabricated values.

\subsection{Failure Mode 3 Statelessness: Ignoring the Provided Before-State}
\label{app:halluc_statelessness}

In this failure mode, the model contradicts the \emph{explicitly provided} before-config. Because our constraint-driven paradigm supplies the full before-state in every prompt, this failure cannot be attributed to implicit state tracking; the model overrides observed facts with a hallucinated prior.

\paragraph{Sample: \texttt{complete\_task} ($|\Delta|=2$, DeepSeek-V3.2).}
The tool call marks task \texttt{T-002} as completed. The before-config provided to the model explicitly states:

\begin{quote}
\small
\texttt{tasks.T-002.status}: \texttt{"in progress"} \\
\texttt{tasks.T-002.completed\_at}: \texttt{null}
\end{quote}

The executor therefore accepts the call: the task transitions from \texttt{"in progress"} to \texttt{"completed"}, and \texttt{completed\_at} is set to the current epoch.

\begin{table}[htbp]
\centering
\small
\setlength{\tabcolsep}{4pt}
\caption{%
  Statelessness: the model asserts the task is ``already completed'' despite
  the before-config explicitly showing \texttt{status = "in progress"}.
}
\label{tab:halluc_stateless}
\begin{tabular}{p{2.3cm} p{4.3cm} p{4.3cm}}
\toprule
& \textbf{Ground truth} & \textbf{DeepSeek-V3.2 prediction} \\
\midrule
\textbf{Before-state} \newline (given in prompt) &
  \texttt{T-002.status}: \texttt{"in progress"} \newline
  \texttt{T-002.completed\_at}: \texttt{null} &
  \emph{(same prompt)} \\
\addlinespace
\textbf{Feedback} &
  \texttt{"Task 'T-002' marked as completed."} &
  \texttt{"error: Task 'T-002' is \textbf{already completed}"} \\
\addlinespace
\textbf{State changes} &
  \texttt{T-002.status}: \texttt{"in progress"} $\to$ \texttt{"completed"} \newline
  \texttt{T-002.completed\_at}: \texttt{null} $\to$ \texttt{1774080132.66} &
  \emph{(no changes predicted)} \\
\bottomrule
\end{tabular}
\end{table}

The model asserts \texttt{"already completed"} while simultaneously having access to the before-config that says the opposite. This is not a state-tracking failure arising from implicit multi-turn inference. It explicitly contradicts the data present in the prompt for a single-turn prediction. The most likely cause is that the model's
parametric knowledge about task-management workflows (where double-completion calls are commonly guarded against) overrides its attention to the provided state. This pattern represents the purest form of hallucination in the environment simulation setting: the model's world model supersedes the ground truth it was given. The same sample was also misclassified by Gemini-3.1-Pro-Preview (which hallucinated a ``dependency not met'' constraint) and MiniMax-M2.7 (which hallucinated a ``task not found'' error), each substituting a different invented state for the explicitly provided one.

\section{Cross-Model Capability Profile Analysis}
\label{app:capability_profiles}

\paragraph{GPT-5.4 vs.\ Gemini-3.1-Pro-Preview: same aggregate CM, different per-delta profiles.}
Both models achieve 42.0\% overall CM, but the per-delta breakdown from Appendix~\ref{app:detailed_results} reveals a crossover structure: Gemini-3.1-Pro-Preview leads at $\Delta\in\{1,2\}$ (68\%/28\% vs.\ 60\%/20\% for GPT-5.4); both are identical at $\Delta\in\{3,4\}$ (36\%/24\%); GPT-5.4 leads substantially at $\Delta\in\{7,8\}$ (44\%/25\% vs.\ 31\%/12\%). This suggests that Gemini-3.1-Pro-Preview has stronger alignment for precise, low-complexity field prediction, while GPT-5.4 has higher resilience against the most complex multi-branch conditional logic. For practitioners choosing a simulation backend, Gemini-3.1-Pro-Preview may be preferable for environments dominated by simple CRUD-style operations, while GPT-5.4 may be preferable for environments with deeply nested state transitions.

\paragraph{DeepSeek-V3.2: strong FM, weak CM.}
DeepSeek-V3.2 achieves 72.5\% overall FM (fourth highest) but only 32.5\% CM (lowest), a 40~pp gap that is the largest FM--CM divergence in the evaluation. On state-changing samples, the gap reaches 58.7~pp (68.7\% FM vs.\ 10.0\% CM). As Table~\ref{tab:fm_t_cm_f} shows, 75--87\% of its CM failures are accompanied by FM\,=\, True across delta levels 1--5: the model generates correctly formatted, domain-appropriate response strings while substantially underperforming on code-execution tracing. This makes it a high-risk choice as a simulation backend for generating training data.

\paragraph{FM--CM gap as a model-selection signal.}
Table~\ref{tab:fm_cm_gap} ranks all models by the FM--CM gap on state-changing samples. A large positive gap indicates the model generates believable but inaccurate simulations; a near-zero or negative gap indicates visible failures. For use cases where the reliability of the training signal matters more than surface fluency, CM alone is the appropriate criterion for backend selection.

\begin{table}[htbp]
\centering
\small
\caption{FM--CM gap on state-changing samples. The gap is the difference between marginal feedback and configuration accuracy; joint error rates are reported separately.
  \textbf{Bold}: column-wise maximum;
  \textcolor{red}{red}: column-wise minimum.}
\label{tab:fm_cm_gap}
\begin{tabular}{lccc}
\toprule
\textbf{Model} & \textbf{FM (state-chg)} & \textbf{CM (state-chg)} & \textbf{Gap (FM\,$-$\,CM)} \\
\midrule
GLM-5
  & \textbf{81.7\%} & 21.3\%          & \textbf{60.4~pp} \\
DeepSeek-V3.2
  & 68.7\%          & \textcolor{red}{10.0\%} & 58.7~pp \\
GPT-5.4
  & 75.7\%          & 22.7\%          & 53.0~pp \\
Gemini-3.1-Pro-Preview
  & 72.7\%          & 22.7\%          & 50.0~pp \\
Qwen3.5-397B-A17B
  & 64.3\%          & \textbf{23.0\%} & 41.3~pp \\
MiniMax-M2.7
  & 21.3\%          & 22.7\%          & $-$1.4~pp \\
Claude-Sonnet-4.6
  & \textcolor{red}{12.0\%} & 17.3\% & \textcolor{red}{$-$5.3~pp} \\
\bottomrule
\end{tabular}
\end{table}
\section{Experimental Setup Details}
\label{app:eval_setup}

\subsection{Frontier LLM Evaluation}
All seven frontier models are queried via their respective inference APIs in non-thinking mode. Every model receives identical prompts instantiating the MDP triple $(s_t, a_t, \texttt{code}(a_t))$; no model has access to prior turns or trajectory history. Ground-truth labels $(o_t, s'_t)$ are produced entirely by the deterministic external executor, making evaluation LLM-free and immune to circular validation.
\subsection{Model Identifiers and API Endpoints}
\label{app:model_ids}
 
Table~\ref{tab:model_ids} maps each display name used throughout the paper to the
exact API model string queried during evaluation, together with the model provider.
All models were called via their respective public inference APIs in
\textbf{non-thinking} (standard) mode with no system-prompt modifications.
 
\begin{table}[htbp]
\centering
\small
\setlength{\tabcolsep}{6pt}
\caption{Display name, exact API model string, and provider for each evaluated frontier model.}
\label{tab:model_ids}
\begin{tabular}{lll}
\toprule
\textbf{Display name (used in paper)} & \textbf{API model string} & \textbf{Provider} \\
\midrule
DeepSeek-V3.2         & \texttt{deepseek-v3.2}              & DeepSeek      \\
Qwen3.5-397B-A17B     & \texttt{qwen3.5-397b-a17b}          & Alibaba Cloud \\
GPT-5.4               & \texttt{gpt-5.4}                    & OpenAI        \\
Gemini-3.1-Pro-Preview        & \texttt{gemini-3.1-pro-preview}     & Google DeepMind \\
Claude-Sonnet-4.6     & \texttt{claude-sonnet-4-6}          & Anthropic     \\
MiniMax-M2.7          & \texttt{MiniMax-M2.7}               & MiniMax       \\
GLM-5                 & \texttt{glm-5}                      & Zhipu AI      \\
\bottomrule
\end{tabular}
\end{table}

\subsection{SFT Training Configuration}
We fine-tune Qwen3-4B-Base using full-parameter SFT via LLaMA-Factory~\citep{zheng-etal-2024-llamafactory} on 2$\times$A800 (80\,GB) GPUs with DeepSpeed ZeRO-3 parallelism~\citep{rajbhandari2020zero}. Key hyperparameters: learning rate $2\times10^{-5}$ with a cosine schedule and 5\% linear warm-up; 3 training epochs; per-device batch size 1 with gradient accumulation over 16 steps (effective batch size 32); bf16 mixed precision with FlashAttention-2. The maximum sequence length is set to 8{,}192 tokens. Each sample encodes the full transition logic $\texttt{code}(a_t)$ together with the structured before-state $s_t$, with token counts peaking near 5{,}000 across the corpus; this window ensures no truncation during training or evaluation. Inference is served via vLLM with a maximum generation length of 16{,}384 tokens and 90\% GPU memory utilization.

All experiments are evaluated under the \texttt{env\_id} holdout protocol, which excludes all 167 benchmark environments from training, including Findings 4 and 5 and the composition ablation in Appendix~\ref{app:composition_ablation}.

\section{Data Composition Ablation}
\label{app:composition_ablation}

Table~\ref{tab:sft_composition} reports the full composition ablation comparing three data mixture strategies, with frontier model performance as a reference.

\textbf{Change-only} (trained exclusively on state-mutating samples) achieves the highest CM on state-changing subsets (60\% Simple, 24\% Medium, 32\% Difficult) but collapses to 0\% FM on Failure and No-Change groups, despite maintaining 96\% CM on these same samples. The model correctly predicts that no state change should occur in most cases, but has never learned to generate the corresponding feedback strings: outputting state-change-style success messages where error messages or minimal acknowledgments are expected. In a deployed pipeline, this feedback corruption directly poisons the agent's reward signal: the environment state is preserved, but the agent receives systematically wrong feedback for every failed or no-op action, causing it to misattribute outcomes and learn incorrect action-consequence associations.

\textbf{Balance} adds failure and no-change samples but does not mirror the empirical $|\Delta|$ distribution, resulting in weaker performance on complex-change tiers. \textbf{Balance2} additionally upweights complex-change samples to match the source distribution, yielding the best overall CM (45.3\%) and FM (79.5\%), surpassing all frontier models on CM.

\begin{table}[htbp]
\centering
\small
\setlength{\tabcolsep}{3.5pt}
\caption{Data composition ablation under \texttt{env\_id} holdout.
\textbf{Bold}: column-wise maximum;
\textcolor{red}{red}: column-wise minimum.}
\label{tab:sft_composition}
\begin{tabular}{l cc cc cc cc cc}
\toprule
& \multicolumn{2}{c}{\textbf{NC/F}}
& \multicolumn{2}{c}{\textbf{Simple}}
& \multicolumn{2}{c}{\textbf{Medium}}
& \multicolumn{2}{c}{\textbf{Difficult}}
& \multicolumn{2}{c}{\textbf{Overall}} \\
\cmidrule(lr){2-3}\cmidrule(lr){4-5}\cmidrule(lr){6-7}\cmidrule(lr){8-9}\cmidrule(lr){10-11}
\textbf{Model} & FM & CM & FM & CM & FM & CM & FM & CM & FM & CM \\
\midrule
Change-only
  & \textcolor{red}{0\%}    & \textcolor{red}{96\%}
  & \textbf{96\%}           & \textbf{60\%}
  & \textbf{91.5\%}         & \textbf{24\%}
  & 52\%                    & \textbf{32\%}
  & \textcolor{red}{64.3\%} & \textbf{47.5\%} \\
Balance
  & 73\%          & 99\%
  & \textbf{96\%} & 54\%
  & 89.0\%        & 23\%
  & 38\%          & \textcolor{red}{6\%}
  & 79.5\%        & 43.8\% \\
\textbf{Balance2 (Ours)}
  & 78\%                  & 99\%
  & 92\%                  & 58\%
  & 89.5\%                & 23.5\%
  & \textcolor{red}{30\%} & 12\%
  & 79.5\%                & 45.3\% \\
\midrule
\multicolumn{11}{l}{\emph{Frontier reference:}} \\
GLM-5 (best FM)
  & 77\%          & \textbf{100\%}
  & 94\%          & \textcolor{red}{44\%}
  & 81.5\%        & \textcolor{red}{14\%}
  & \textbf{70\%} & 28\%
  & \textbf{80.5\%} & \textcolor{red}{41.0\%} \\
Qwen3.5-397B-A17B (best CM)
  & \textbf{83\%}         & \textbf{100\%}
  & \textcolor{red}{84\%} & 46\%
  & \textcolor{red}{65.0\%} & 17\%
  & 42\%                  & 24\%
  & 69.0\%                & 42.3\% \\
\bottomrule
\end{tabular}
\end{table}

\section{Pre-SFT Scaling Baseline}
\label{app:scaling}

Table~\ref{tab:baseline_results} shows Qwen3.5 instruct models (4B, 9B, 27B) without fine-tuning. Clear FM scaling trends emerge (2.3\%$\to$62.8\%), but even the 27B model is substantially below the best-performing LLMs on the frontier, confirming that specialization is required beyond scale alone.

\begin{table}[htbp]
\centering
\small
\caption{Qwen3.5 non-thinking instruct models before SFT.
  \textbf{Bold}: column-wise maximum;
  \textcolor{red}{red}: column-wise minimum.}
\label{tab:baseline_results}
\begin{tabular}{l cc cc cc cc cc}
\toprule
& \multicolumn{2}{c}{NC/F} & \multicolumn{2}{c}{Simple} &
  \multicolumn{2}{c}{Medium} & \multicolumn{2}{c}{Difficult} &
  \multicolumn{2}{c}{Overall} \\
\cmidrule(lr){2-3}\cmidrule(lr){4-5}\cmidrule(lr){6-7}\cmidrule(lr){8-9}\cmidrule(lr){10-11}
Model & FM & CM & FM & CM & FM & CM & FM & CM & FM & CM \\
\midrule
Qwen3.5-4B
  & \textcolor{red}{8\%}  & \textcolor{red}{95\%}
  & \textcolor{red}{0\%}  & \textbf{42\%}
  & \textcolor{red}{0.5\%}  & 10.5\%
  & \textcolor{red}{0\%}  & 6\%
  & \textcolor{red}{2.3\%}  & 35.0\% \\
Qwen3.5-9B
  & 39\% & 98\%
  & 6\%  & \textbf{42\%}
  & 1.5\%  & \textbf{11.6\%}
  & \textcolor{red}{0\%}  & \textbf{20\%}
  & 11.3\% & \textbf{38.1\%} \\
Qwen3.5-27B
  & \textbf{82\%} & \textbf{99\%}
  & \textbf{78\%} & \textcolor{red}{26\%}
  & \textbf{58.5\%} & \textcolor{red}{2.0\%}
  & \textbf{26\%} & \textcolor{red}{4\%}
  & \textbf{62.8\%} & \textcolor{red}{29.5\%} \\
\bottomrule
\end{tabular}
\end{table}


\section{Training Method Comparison}
\label{app:training_method}

Table~\ref{tab:sft_method} compares LoRA and full-parameter fine-tuning on Qwen3-4B-Base at a 5K-sample budget under the  \texttt{env\_id} holdout. All hyperparameters are held fixed as described in Appendix~\ref{app:eval_setup}.

\begin{table}[htbp]
\centering
\small
\setlength{\tabcolsep}{3.5pt}
\caption{Training method comparison on Qwen3-4B-Base (5K samples,
\texttt{env\_id} holdout). Full fine-tuning outperforms LoRA on CM,
particularly in the Simple tier.
  \textbf{Bold}: column-wise maximum;
  \textcolor{red}{red}: column-wise minimum.}
\label{tab:sft_method}
\begin{tabular}{l cc cc cc cc cc}
\toprule
& \multicolumn{2}{c}{\textbf{NC/F}} & \multicolumn{2}{c}{\textbf{Simple}} &
  \multicolumn{2}{c}{\textbf{Medium}} & \multicolumn{2}{c}{\textbf{Difficult}} &
  \multicolumn{2}{c}{\textbf{Overall}} \\
\cmidrule(lr){2-3}\cmidrule(lr){4-5}\cmidrule(lr){6-7}\cmidrule(lr){8-9}\cmidrule(lr){10-11}
\textbf{Model} & FM & CM & FM & CM & FM & CM & FM & CM & FM & CM \\
\midrule
Before SFT
  & \textcolor{red}{18\%} & \textcolor{red}{51\%}
  & \textcolor{red}{34\%} & \textbf{40\%}
  & \textcolor{red}{34\%}   & \textbf{22.5\%}
  & \textcolor{red}{10\%} & \textcolor{red}{4\%}
  & \textcolor{red}{27.0\%} & \textcolor{red}{29.5\%} \\
LoRA-5K (w/o reason)
  & \textbf{67\%} & \textbf{99\%}
  & \textbf{90\%} & \textcolor{red}{20\%}
  & 78.9\% & 5.0\%
  & \textbf{46\%} & 16\%
  & \textbf{73.2\%} & 31.8\% \\
LoRA-5K (w/ reason)
  & 66\% & \textbf{99\%}
  & \textbf{90\%} & 26\%
  & 77.5\% & 6.0\%
  & 34\% & \textbf{20\%}
  & 70.8\% & 33.5\% \\
Full-5K (w/o reason)
  & 65\% & \textbf{99\%}
  & 86\% & 34\%
  & \textbf{84.0\%} & 10.5\%
  & 30\% & 6\%
  & 72.8\% & \textbf{35.0\%} \\
Full-5K (w/ reason)
  & 63\% & \textbf{99\%}
  & 88\% & \textcolor{red}{20\%}
  & 80.5\% & \textcolor{red}{3.5\%}
  & 32\% & 10\%
  & 71.0\% & 30.3\% \\
\bottomrule
\end{tabular}
\end{table}

Full-parameter SFT raises FM from 27\% to 72.8\% ($2.7\times$) and aligns the model's output schema with the expected response convention, correcting near-zero FM on Failure and No-Change samples (CM rises from 51\% to 99\% on these groups). For CM, the state-change cliff persists at the 5K-sample scale: Medium-group CM reaches only 10.5\%, and Difficult-group CM falls to 6\%, confirming that the cliff reflects a fundamental capability gap rather than a formatting artifact.

Full-parameter fine-tuning outperforms LoRA on CM under the same training-data budget (35.0\% vs.\ 31.8\%), with the advantage concentrated in the Simple tier (34\% vs.\ 20\%). We attribute this to LoRA's limited representational capacity, which overwrites the dense attention patterns that underlie sequential code-execution tracing. The state-change cliff persists under both methods at 5K samples, confirming it as a genuine capability gap rather than a parameter-efficiency artifact.

\section{Supplementary Reasoning-Mode Evaluation}
\label{app:reasoning_eval}

We additionally evaluate the seven frontier models with reasoning enabled.
Table~\ref{tab:reasoning_baselines} reports feedback and configuration accuracy
for the state-preserving and state-changing groups, together with overall results.
This evaluation concerns inference-time reasoning and is distinct from the use
of reasoning traces as SFT targets in Appendix~\ref{app:training_method}.

For compact presentation, DeepSeek denotes DeepSeek-V3.2; Qwen denotes
Qwen3.5-397B-A17B; GPT denotes GPT-5.4; Gemini denotes Gemini-3.1-Pro-Preview;
Claude denotes Claude-Sonnet-4.6; MiniMax denotes MiniMax-M2.7; and GLM denotes
GLM-5. These abbreviations are also used in Appendix~\ref{app:expanded_eval}.

\begin{table}[tbp]
\centering
\small
\setlength{\tabcolsep}{4pt}
\caption{Supplementary reasoning-mode results (\%). Failure + No-Change
denotes state-preserving samples. Overall results use sample weighting.}
\label{tab:reasoning_baselines}
\begin{tabular}{lrrrrrr}
\toprule
& \multicolumn{2}{c}{Failure + No-Change}
& \multicolumn{2}{c}{State-Change}
& \multicolumn{2}{c}{Overall} \\
\cmidrule(lr){2-3}\cmidrule(lr){4-5}\cmidrule(lr){6-7}
Model & FM & CM & FM & CM & FM & CM \\
\midrule
DeepSeek & 79.00 & 99.00 & 65.67 & 19.00 & 69.00 & 39.00 \\
Qwen & 80.00 & 100.00 & 69.67 & 24.67 & 72.25 & 43.50 \\
GPT & 67.00 & 100.00 & 74.00 & 22.67 & 72.25 & 42.00 \\
Gemini & 77.00 & 100.00 & 86.33 & 24.00 & 84.00 & 43.00 \\
Claude & 79.00 & 100.00 & 60.67 & 19.67 & 65.25 & 39.75 \\
MiniMax & 73.00 & 100.00 & 20.33 & 22.33 & 33.50 & 41.75 \\
GLM & 75.00 & 99.00 & 80.67 & 22.00 & 79.25 & 41.25 \\
\bottomrule
\end{tabular}
\end{table}

Reasoning-mode overall CM ranges from 39.00\% to 43.50\%, with Qwen obtaining
the highest score. Gemini achieves the highest overall FM at 84.00\%.
State-preserving CM remains at 99--100\%, whereas state-changing CM ranges
from 19.00\% to 24.67\%. Thus, substantial state-transition errors remain
under reasoning-enabled inference. The aggregate results do not determine
whether reasoning helps or hurts at each individual change count.

\section{Expanded Evaluation by State-Change Count}
\label{app:expanded_eval}

\subsection{Expanded sample set and results}

We synthesize an expanded set with 100 samples at each state-change count
$k=|\Delta|\in\{0,\ldots,12\}$, for 1{,}300 samples in total.
Tables~\ref{tab:expanded_fm} and~\ref{tab:expanded_cm} report FM and CM
separately for all seven models. This evaluation is reported separately from
the original 400-sample benchmark because its sampling distribution differs.
The $k=0$ group contains transitions with no recorded configuration change;
the results below are stratified by change count rather than action outcome.

\begin{table}[tbp]
\centering
\small
\setlength{\tabcolsep}{4pt}
\caption{Expanded evaluation: FM (\%) by state-change count.
Each row contains 100 samples. Model abbreviations are defined in
Appendix~\ref{app:reasoning_eval}.}
\label{tab:expanded_fm}
\begin{tabular}{rrrrrrrr}
\toprule
$k$ & DeepSeek & Qwen & GPT & Gemini & Claude & MiniMax & GLM \\
\midrule
0 & 68 & 69 & 57 & 61 & 64 & 62 & 62 \\
1 & 87 & 93 & 78 & 95 & 66 & 27 & 95 \\
2 & 88 & 92 & 69 & 90 & 70 & 33 & 93 \\
3 & 100 & 84 & 63 & 96 & 80 & 24 & 91 \\
4 & 78 & 71 & 64 & 89 & 67 & 27 & 82 \\
5 & 85 & 68 & 48 & 87 & 65 & 27 & 85 \\
6 & 71 & 69 & 55 & 82 & 76 & 8 & 75 \\
7 & 28 & 78 & 55 & 89 & 67 & 11 & 78 \\
8 & 25 & 38 & 37 & 64 & 75 & 0 & 75 \\
9 & 20 & 0 & 20 & 48 & 60 & 0 & 40 \\
10 & 25 & 42 & 50 & 52 & 56 & 17 & 50 \\
11 & 0 & 33 & 33 & 48 & 52 & 16 & 50 \\
12 & 40 & 42 & 40 & 45 & 50 & 23 & 56 \\
\bottomrule
\end{tabular}
\end{table}

\begin{table}[tbp]
\centering
\small
\setlength{\tabcolsep}{4pt}
\caption{Expanded evaluation: CM (\%) by state-change count.
Each row contains 100 samples. Model abbreviations are defined in
Appendix~\ref{app:reasoning_eval}.}
\label{tab:expanded_cm}
\begin{tabular}{rrrrrrrr}
\toprule
$k$ & DeepSeek & Qwen & GPT & Gemini & Claude & MiniMax & GLM \\
\midrule
0 & 100 & 100 & 98 & 100 & 100 & 99 & 100 \\
1 & 53 & 73 & 65 & 81 & 65 & 65 & 71 \\
2 & 22 & 38 & 37 & 39 & 36 & 35 & 36 \\
3 & 0 & 29 & 23 & 30 & 24 & 27 & 27 \\
4 & 11 & 19 & 13 & 17 & 18 & 18 & 15 \\
5 & 1 & 6 & 6 & 5 & 3 & 2 & 5 \\
6 & 3 & 5 & 6 & 6 & 6 & 9 & 5 \\
7 & 6 & 19 & 24 & 24 & 16 & 19 & 24 \\
8 & 0 & 4 & 4 & 5 & 7 & 0 & 4 \\
9 & 0 & 3 & 6 & 3 & 2 & 0 & 0 \\
10 & 1 & 3 & 6 & 7 & 7 & 2 & 3 \\
11 & 0 & 7 & 6 & 7 & 8 & 0 & 7 \\
12 & 0 & 2 & 10 & 8 & 7 & 6 & 2 \\
\bottomrule
\end{tabular}
\end{table}

CM ranges from 53\% to 81\% at $k=1$, falls to 22--39\% at $k=2$,
and reaches 1--6\% at $k=5$. The trend is non-monotonic: at $k=7$,
CM ranges from 6\% to 24\%. These results support a broad decline in configuration accuracy for multi-field transitions. Equalizing the sample count
across $k$ reduces the influence of individual examples within each stratum,
but does not by itself separate the effects of change count, update pattern,
tool logic, and runtime-dependent values.

\subsection{Representative bootstrap confidence intervals}
\label{app:expanded_uncertainty}

For each reported model and change count, we resample instances with
replacement within that stratum and compute CM for each resample. The 2.5th
and 97.5th percentiles give the reported 95\% bootstrap interval.
Table~\ref{tab:expanded_bootstrap} presents the representative intervals
available for two models and four change counts. These intervals describe
instance-level resampling uncertainty within the evaluated strata; they do
not measure variation across independently generated datasets or repeated
model runs.

\begin{table}[tbp]
\centering
\small
\setlength{\tabcolsep}{6pt}
\caption{Representative 95\% percentile-bootstrap intervals for CM on the
expanded set. Each row uses $n=100$ instances. Values are percentages.}
\label{tab:expanded_bootstrap}
\begin{tabular}{lrrr}
\toprule
Model & $k$ & CM & 95\% interval \\
\midrule
DeepSeek & 1 & 53.00 & [43.00, 63.00] \\
DeepSeek & 2 & 22.00 & [14.00, 31.00] \\
DeepSeek & 5 & 1.00 & [0.00, 4.00] \\
DeepSeek & 12 & 0.00 & [0.00, 0.00] \\
Gemini & 1 & 81.00 & [72.00, 88.00] \\
Gemini & 2 & 39.00 & [30.00, 49.00] \\
Gemini & 5 & 5.00 & [1.80, 10.20] \\
Gemini & 12 & 8.00 & [3.10, 14.50] \\
\bottomrule
\end{tabular}
\end{table}

For DeepSeek at $k=12$, all observed outcomes are failures. Resampling this
all-zero outcome vector produces the degenerate empirical interval $[0,0]$.
This interval does not establish that the population success probability is
zero. The intervals otherwise illustrate uncertainty in the observed CM
rates and should be read alongside the point estimates and sample counts.

\end{document}